\newcommand{\sarathi}{Sarathi\xspace}
\newcommand{\sysname}{Sarathi-Serve\xspace}

\newcommand{\prefill}{\textit{prefill}\xspace}

\newcommand{\decode}{\textit{decode}\xspace}
\newcommand{\chunking}{\textit{chunked-prefills}\xspace}
\newcommand{\Chunking}{\textit{Chunked-prefills}\xspace}

\newcommand{\dcoel}{\textit{hybrid-batching}\xspace}
\newcommand{\hbatch}{\textit{stall-free batching}\xspace}
\newcommand{\Hbatch}{\textit{Stall-free batching}\xspace}

\newcommand{\sharegpt}{\textit{openchat\_sharegpt4}\xspace}
\newcommand{\arxivsum}{\textit{arxiv\_summarization}\xspace}
\newcommand{\relaxed}{\textit{relaxed}\xspace}
\newcommand{\strict}{\textit{strict}\xspace}
\newcommand{\slo}{SLO\xspace}
\newcommand{\slos}{SLOs\xspace}

\newcommand{\myx}{$\times$\xspace}
\newcommand{\llama}{LLaMA\xspace}

\newcommand{\ils}{iteration-level batching\xspace}
\newcommand{\Ils}{Iteration-level batching\xspace}
\newcommand{\rls}{request-level batching\xspace}
\newcommand{\Rls}{Request-level batching\xspace}
\newcommand{\prefillprio}{\textit{prefill-prioritizing}\xspace}
\newcommand{\decodeprio}{\textit{decode-prioritizing}\xspace}

\newcommand{\mistral}{Mistral-7B\xspace}
\newcommand{\llamaL}{LLaMA2-70B\xspace}

\newcommand{\yi}{Yi-34B\xspace}
\newcommand{\falcon}{Falcon-180B\xspace}

\newcommand{\ie}{\textit{i.e.,}\xspace}
\newcommand{\eg}{\textit{e.g.,}\xspace}

\newcommand{\sref}[1]{\S\ref{#1}}

\newcommand{\attn}{\textit{attention}\xspace}

\documentclass[letterpaper,twocolumn,10pt]{article}
\usepackage{usenix2019_v3}

\usepackage{tikz}
\usepackage{amsmath}

\usepackage{filecontents}

\usepackage{multirow}
\usepackage{multicol}
\usepackage{enumitem}
\usepackage{algorithm}
\usepackage{algpseudocode}
\usepackage{subcaption}
\usepackage{titling}
\algtext*{EndWhile} %
\algtext*{EndIf} %
\algtext*{EndFor} %

\usepackage{url}

\usepackage{breakurl}
\usepackage{hyperref}
\makeatletter
\renewcommand{\sectionautorefname}{\S\@gobble}
\renewcommand{\subsectionautorefname}{\S\@gobble}
\renewcommand{\subsubsectionautorefname}{\S\@gobble}
\makeatother

\usepackage{ifthen}
\newboolean{publicversion}
\setboolean{publicversion}{true}

\ifthenelse{\boolean{publicversion}}{
	\newcommand{\grumbler}[3]{}
        \newcommand{\jm}[1]{}
        \newcommand{\ap}[1]{}
        \newcommand{\nk}[1]{}
        \newcommand{\rr}[1]{}
        \newcommand{\amey}[1]{}
        \newcommand{\alexey}[1]{}
}
{
\newcommand{\grumbler}[3]{\xspace\textcolor{#3}{\bf #1: #2}}
\newcommand{\jm}[1]{\grumbler{Jayashree}{#1}{magenta}}
\newcommand{\ap}[1]{\grumbler{Ashish}{#1}{violet}}

\newcommand{\nk}[1]{\grumbler{Nipun}{#1}{teal}}
\newcommand{\rr}[1]{\grumbler{Ram}{#1}{cyan}}

\newcommand{\amey}[1]{\grumbler{Amey}{#1}{green}}
\newcommand{\alexey}[1]{\grumbler{AT}{#1}{green}}
}

\definecolor{dgreen}{rgb}{0.0, 0.5, 0.0}
\usepackage{comment}
\usepackage{xspace}
\usepackage{authblk}
\usepackage{soul}
\usepackage{tikz}
\usepackage{xcolor}
\usepackage{array}
\usepackage{cleveref}
\usepackage{xparse}

\usepackage{booktabs}
\usepackage{cleveref}%
\crefname{section}{\S}{\SS}
\crefformat{section}{\S#2#1#3} %
\crefformat{subsection}{\S#2#1#3}
\crefformat{subsubsection}{\S#2#1#3}

\crefformat{subfigure}{#2\onlyletter{#1}#3}
\crefrangeformat{subfigure}{(#3\onlyletter{#1}#4--#5\onlyletter{#2}#6)}
\crefmultiformat{subfigure}
  {(#2\onlyletter{#1}#3}
  {,~#2\onlyletter{#1}#3)}
  {,~#2\onlyletter{#1}#3)}
  {,~#2\onlyletter{#1}#3)}

\ExplSyntaxOn
\NewDocumentCommand{\onlyletter}{m}
 {
  \tl_set:Nx \l_tmpa_tl { #1 }
  \tl_item:Nn \l_tmpa_tl { -1 }
 }
\ExplSyntaxOff
\date{}
\usepackage[available,functional,reproduced]{usenixbadges}

\title{\Large \bf Taming Throughput-Latency Tradeoff in LLM Inference with \textit{\sysname}}
\begin{document}
\author[2]{Amey Agrawal\thanks{Part of this work was done during an internship at MSR India.}}\author[1]{Nitin Kedia}\author[1]{Ashish Panwar}\author[1]{Jayashree Mohan}\author[1]{Nipun Kwatra}\author[1]{\\Bhargav S. Gulavani}\author[2]{Alexey Tumanov}\author[1]{Ramachandran Ramjee} \affil[1]{Microsoft Research India}\affil[2]{Georgia Institute of Technology}

\maketitle

    \begin{abstract}

Each LLM serving request goes through two phases. The first is \textit{prefill} which processes the entire input prompt and produces the first output token and the second is \textit{decode} which generates the rest of output tokens, one-at-a-time. Prefill iterations have high latency but saturate GPU compute due to parallel processing of the input prompt. In contrast, decode iterations have low latency but also low compute utilization because a decode iteration processes only a single token per request. This makes batching highly effective for decodes and consequently for overall throughput. However, batching multiple requests leads to an interleaving of prefill and decode iterations which makes it challenging to achieve both high throughput and low latency.

We introduce an efficient LLM inference scheduler, \sysname, to address this throughput-latency tradeoff. \sysname introduces \chunking which splits a prefill request into near equal sized chunks and creates \textit{stall-free} schedules that adds new requests in a batch without pausing ongoing decodes. Stall-free scheduling unlocks the opportunity to improve throughput with large batch sizes while minimizing the effect of batching on latency. Furthermore, uniform batches in \sysname ameliorate the imbalance between iterations, resulting in minimal pipeline bubbles. 

Our techniques yield significant improvements in inference performance across models and hardware under tail latency constraints. For Mistral-7B on single A100 GPUs, we achieve 2.6\myx higher serving capacity and up to 3.7\myx higher serving capacity for the Yi-34B model on two A100 GPUs as compared to vLLM. When used with pipeline parallelism on Falcon-180B, \sysname provides up to 5.6\myx gain in the end-to-end serving capacity. The source code for \sysname is available at~\href{https://github.com/microsoft/sarathi-serve}{https://github.com/microsoft/sarathi-serve}.

\if
We present \sysname to address these challenges. \sysname employs \chunking,  which splits a prefill request into near equal sized chunks, and \hbatch,  which coalesces the on-going decodes with prefill chunk(s) from new requests such that each batch has a uniform, predetermined number of tokens.
During inference, the prefill chunk saturates GPU compute, while the decode requests `piggyback' and cost up to an order of magnitude less compared to a decode-only batch. Furthermore, the uniform compute design of these batches ameliorates the imbalance between iterations, resulting in significantly reduced tail latency and minimal pipeline bubbles.

Our techniques yield significant improvements in inference performance across models and hardware under tail latency constraints. For the Yi-34B model on two NVIDIA A100 GPU, \sysname improves the system serving capacity by up to 2.8\myx under different tail latency compared to vLLM \slos. For Mistral-7B on single A100 GPUs, we achieve 2.3\myx higher serving capacity. When used with pipeline parallelism on Falcon-180B, \sysname provides up to an order of magnitude gain in the end-to-end serving capacity. \sysname's source code will be made publicly available to facilitate open-source development of LLM inference systems.
\fi

\end{abstract}

\section{Introduction}\label{sec:Introduction}

Large language models (LLMs)~\cite{wei2022emergent,gpt3-brown2020language,chowdhery2022Palm,openai2022gpt4techreport,kaplan2020scalinglaws} have shown impressive abilities in a wide variety of tasks spanning natural language processing, question answering, code generation, etc. This has led to tremendous increase in their usage across many applications such as chatbots~\cite{openai2022gpt4techreport,chatgpt,claudeai,characterai}, search~\cite{bingai,komoai,youdotcom,perplexityai,bard}, code assistants~\cite{githubcopilot,replitghostwriter,amazoncodewhisperer}, etc. The significant GPU compute required for running inference on large models, coupled with significant increase in their usage, has made LLM inference a dominant GPU workload today. Thus, optimizing LLM inference has been a key focus for many recent systems~\cite{efficiently-scaling-transformer-inference,flexgen,orca,vllmsosp,patel2023splitwise,distserve2024,sarathi2023}.

\begin{figure}
    \begin{subfigure}[b]{0.235\textwidth}        
        \includegraphics[ width=\textwidth]{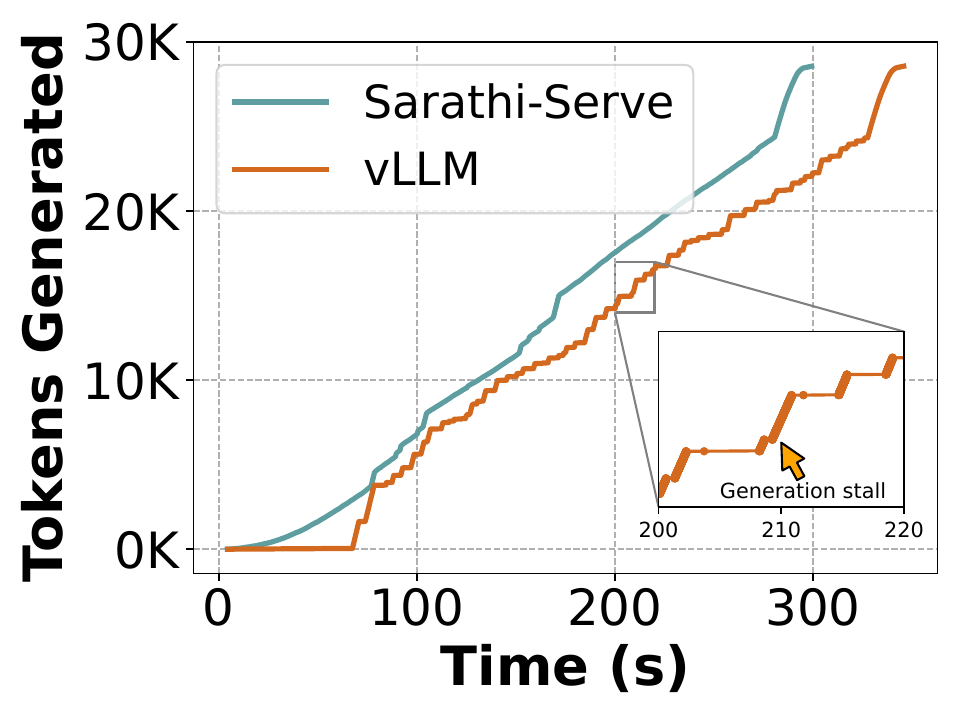}
        \caption{Generation stall.}
        \label{fig:banner:stal}
    \end{subfigure}
    \begin{subfigure}[b]{0.23\textwidth}
        \includegraphics[width=\textwidth]{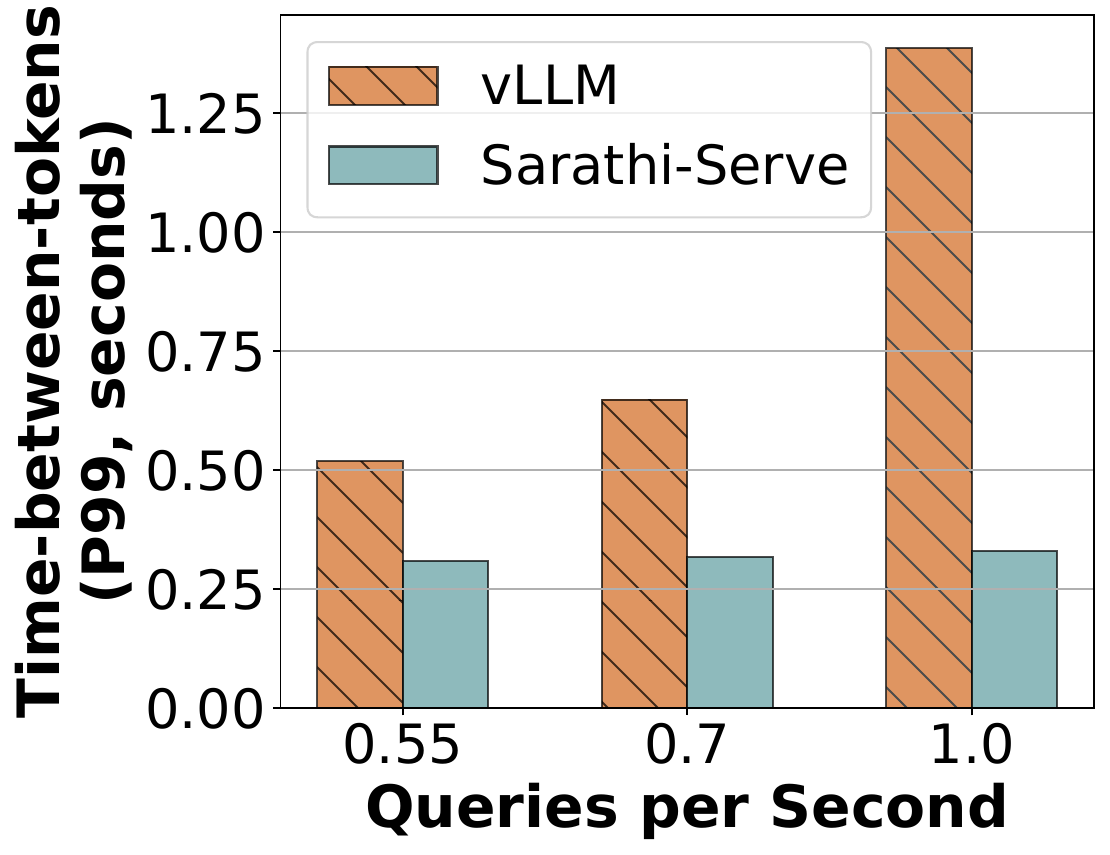}
        \caption{High tail latency.}
        \label{fig:banner:latency}
    \end{subfigure}
    \caption{
    \yi running on two A100 GPUs serving 128 requests from \textit{arxiv-summarisation} trace. ~\ref{fig:banner:stal} highlights one of the many generation stalls lasting over several seconds in vLLM ~\cite{vllmsosp}. ~\ref{fig:banner:latency} shows the impact of increasing load on tail latency. \sysname improves throughput while eliminating generation stalls.}
    \vspace{-1em}
\end{figure}

Optimizing throughput and latency are both important objectives in LLM inference since the former helps keep 
serving costs tractable while the latter is necessary to meet application requirements. In this paper, we show that current LLM serving systems have to face a tradeoff between throughput and latency. In particular, LLM inference throughput can be increased significantly with batching. However, the way existing systems batch multiple requests leads to a compromise on either throughput or latency. For example,~\autoref{fig:banner:latency} shows that increasing load can significantly increase tail latency in a state-of-the-art LLM serving system vLLM ~\cite{vllmsosp}.

Each LLM inference request goes through two phases -- a \textit{prefill} phase followed by a \textit{decode} phase. The \textit{prefill} phase corresponds to the processing of the input prompt and the \textit{decode} phase corresponds to the autoregressive token generation. The prefill phase is compute-bound because it processes all tokens of an input prompt in parallel whereas the decode phase is memory-bound because it processes only one token per-request at a time. Therefore, decodes benefit significantly from batching because larger batches can use GPUs more efficiently whereas prefills do not benefit from batching.

\begin{figure}[t!]
    \centering
        \includegraphics[width=0.4\textwidth]{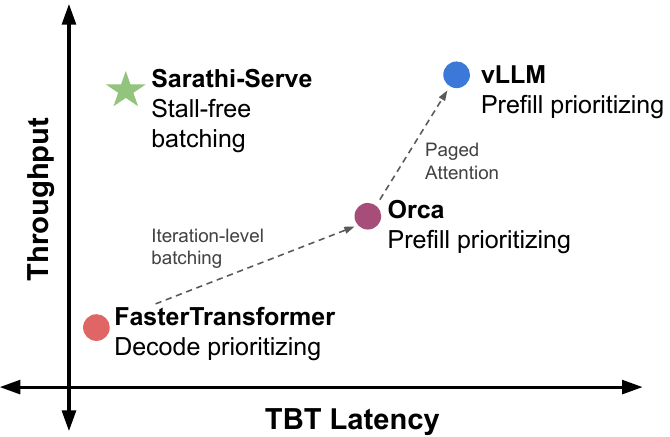}

    \caption{ Current LLM serving systems involve a tradeoff between throughput and latency depending on their scheduling policy. Prioritizing prefills optimizes throughput but sacrifices TBT (time-between-tokens) tail latency whereas prioritizing decodes has the opposite effect. \sysname serves high throughput with low TBT latency via stall-free batching. (The figure is illustrative and actual values will depend on the model and workload characteristics.)}
       \label{fig:intro:tradeoff-space}
\end{figure}

Current LLM inference schedulers can be broadly classified into two categories\footnote{We classify recent schedulers Splitwise~\cite{patel2023splitwise} and DistServe~\cite{distserve2024} under a third category ``disaggregated'' and discuss them in~\autoref{section:relatedwork}.}, namely, 
\prefillprio and \decodeprio depending on how they schedule the prefill and decode phases while batching requests. In this paper, we argue that  both strategies have fundamental pitfalls that make them unsuitable for serving online inference (see~\autoref{fig:intro:tradeoff-space}).

Traditional request-level batching systems such as FasterTransformer~\cite{fastertransformer} employ \decodeprio scheduling. These systems submit a batch of requests to the execution engine that first computes the prefill phase of all requests and then schedules their decode phase. The batch completes only after all requests in it have finished their decode phase i.e., new prefills are not scheduled as long as one or more requests are doing decodes. This strategy optimizes inference for latency metric time-between-tokens or TBT -- an important performance metric for LLMs. This is because new requests do not affect the execution of ongoing requests in their  decode phase. However, \decodeprio schedulers severely compromise on throughput because even if some requests in a batch finish early, the execution continues with reduced batch size until the completion of the last request.

Orca~\cite{orca} introduced iteration-level batching wherein requests can dynamically enter or exit a batch at the granularity of individual iterations. Iteration-level batching improves throughput by avoiding inefficiencies of request-level batching systems. Orca and several other recent systems like vLLM~\cite{vLLM:github} combine iteration-level batching with \prefillprio scheduling wherein they eagerly schedule the prefill phase of one or more requests first i.e., whenever GPU memory becomes available. This way, \prefillprio schedulers have better throughput because computing prefills first allows subsequent decodes to operate at high batch sizes. However, prioritizing prefills leads to high latency because it interferes with ongoing decodes. Since prefills can take arbitrarily long time depending on the lengths of the given prompts, \prefillprio schedulers lead to an undesirable phenomenon that we refer to as \textit{generation stalls} in this paper. For example, ~\autoref{fig:banner:stal} shows that a generation stall in vLLM can last over several seconds.

Another challenge introduced by traditional iteration-level scheduling systems like Orca~\cite{orca} is pipeline stalls or bubbles~\cite{gpipe}. These appear in pipeline-parallelism (PP) deployments that are needed to scale LLM inference across several nodes. In servers with high bandwidth connectivity such as NVIDIA DGX A100~\cite{nvidiadgx}, tensor-parallelism (TP)~\cite{megatron} can enable deployment of an LLM on up to 8 GPUs,  supporting large batch sizes with low latencies. However, TP can have prohibitively high latencies when hyper-clusters are unavailable~\cite{varuna}. Thus, as an alternative to TP, pipeline-parallelism (PP)~\cite{pipedream, varuna} is typically used across commodity networks. Existing systems rely on micro-batches to mitigate pipeline stalls or bubbles~\cite{gpipe}. However, the standard micro-batch based scheduling can still lead to pipeline bubbles due to the unique characteristics of LLM inference. Specifically, LLM inference consists of a mixture of varying length prefills and decodes. The resulting schedule can thus have wildly varying runtimes across different micro-batches that waste GPU cycles and degrade the overall system throughput.

To address these challenges, we propose \sysname, a scheduler to balance the throughput-latency tradeoff for scalable online LLM inference serving. \sysname is based on two key ideas: \chunking and \textit{stall-free} scheduling. \Chunking splits a prefill request into equal
compute-sized chunks and computes a prompt's prefill phase over multiple iterations (each with a subset of the prompt tokens). \textit{Stall-free} scheduling allows \textit{new requests to join a running batch without pausing ongoing decodes}. This  involves constructing a batch by coalescing all the on-going decodes with one (or more) prefill chunks from new requests such that each batch reaches the pre-configured chunk size. \sysname builds upon iteration-level batching but with an important distinction: it throttles the number of prefill tokens in each iteration while admitting new requests in a running batch. This not only bounds the latency of each iteration, but also makes it nearly independent of the total length of input prompts. This way, \sysname minimizes the effect of computing new prefills on the TBT of ongoing decodes enabling both high throughput and low TBT latency. %

In addition, hybrid batches (consisting of prefill and decode tokens) constructed by \sysname have a near-uniform compute requirement. With pipeline-parallelism, this allows us to create balanced micro-batching based schedules that significantly reduce pipeline bubbles and improve GPU utilization, thus allowing efficient and scalable deployments.

We evaluate \sysname across different models and hardware --- \mistral on a single A100, \yi on 2 A100 GPUs with 2-way tensor parallelism, \llamaL on 8 A40 GPUs, and \falcon with 2-way pipeline and 4-way tensor parallelism across 8 A100 GPUs connected over commodity ethernet. For \yi, \sysname improves system serving capacity by up to 3.7\myx under different \slo targets. Similarly for \mistral, we achieve up to 2.6\myx higher serving capacity. 
\sysname also reduces pipeline bubbles, resulting in up to 5.6\myx gains in end-to-end serving capacity for Falcon-180B deployed with pipeline parallelism.

The main contributions of our paper include:
\begin{enumerate}[noitemsep,topsep=0pt,parsep=0pt,partopsep=0pt]
    \item We identify a number of pitfalls in the current LLM serving systems, particularly in the context of navigating the throughput-latency tradeoff.
    \item We introduce two simple-yet-effective techniques, \chunking and \hbatch, to improve the performance of an LLM serving system. 
    \item We show generality through extensive evaluation over multiple models, hardware, and parallelism strategies demonstrating that \sysname improves model serving capacity by up to an order of magnitude.
\end{enumerate}

\section{Background}
\label{sec:background}

In this section, we describe the typical LLM model architecture along with their auto-regressive inference process. We also provide an overview of the scheduling policies and important performance metrics.

\subsection{The Transformer Architecture}
\label{sec:background:llm-architecure}

Popular large language models, like, GPT-3 \cite{gpt3}, \llama \cite{touvron2023llama}, Yi \cite{yi} etc. are decoder-only transformer models trained on next token prediction tasks. These models consist of a stack of layers identical in structure. Each layer contains two modules -- self-attention and feed-forward network (FFN).

\noindent{\bf Self-attention module:} The self-attention module is central to the transformer architecture~\cite{attentionpaper}, enabling each part of a sequence to consider all previous parts for generating a contextual representation. During the computation of self-attention, first the Query ($Q$), Key ($K$) and Value ($V$) vectors corresponding to each input token are obtained via a linear transformation. Next, the \attn operator computes a semantic relationship among all tokens of a sequence. This involves computing a dot-product of each $Q$ vector with $K$ vectors of all preceding tokens of the sequence, followed by a softmax operation to obtain a weight vector, which is then used to compute a weighted average of the $V$ vectors. This attention computation can be performed across multiple \textit{heads}, whose outputs are combined using a linear transformation. 

\noindent{\bf Feed-forward network (FFN):} FFN typically consists of two linear transformations with a non-linear activation in between. The first linear layer transforms an input token embedding of dimension $h$ to a higher dimension $h2$. This is followed by an activation function, typically ReLU or GELU~\cite{relu2019,gelu2023}. Finally, the second linear layer, transforms the token embedding back to the original dimension $h$.

\subsection{LLM Inference Process}
\label{sec:background:inference-process}

\noindent{\bf Autoregressive decoding:} LLM inference consists of two distinct phases -- a \prefill phase followed by a \decode phase. The prefill phase processes the user's input prompt and produces the first output token. Subsequently, the decode phase generates output tokens one at a time wherein the token generated in the previous step is passed through the model to generate the next token until a special \textit{end-of-sequence} token is generated. Note that the decode phase requires access to all the keys and values associated with all the previously processed tokens to perform the attention operation. To avoid repeated recomputation, contemporary LLM inference systems store activations in KV-cache~\cite{megatron,orca,fastertransformer}.

A typical LLM prompt contains 100s-1000s of input tokens \autoref{table:eval:datasets}, ~\cite{zheng2023lmsyschat1m}. During the prefill phase all these prompt tokens are processed in parallel in a single iteration. The parallel processing allows efficient utilization of GPU compute. On the contrary, the decode phase involves a full forward pass of the model over a single token generated in the previous iteration. This leads to low compute utilization making decodes memory-bound.

\noindent{\bf Batched LLM inference in multi-tenant environment:} A production serving system must deal with concurrent requests from multiple users. Naively processing requests in a sequential manner leads to a severe under-utilization of GPU compute. In order to achieve higher GPU utilization, LLM serving systems leverage batching to process multiple requests concurrently. This is particularly effective for the decode phase processing which has lower computational intensity at low batch sizes. Higher batch sizes allows the cost of fetching model parameters to be amortized across multiple requests.

\if
On the contrary, prefill phase of each request contains sufficient number of token to saturate GPU compute, which renders batching ineffective. \autoref{fig:background:tput} illustrates throughput as a function of batch size, and we can observe that while for decode iterations throughput increases almost linearly with batch size, prefill throughput almost saturates even with a single request.

\noindent{\bf Takeaway-1:} \textit{The two phases of LLM inference -- prefill and decode -- demonstrate contrasting behaviors wherein batching boosts decode phase throughput immensely but has little effect on prefill throughput.}
\fi

Recently, several complementary techniques have been proposed to optimize throughput by enabling support for larger batch sizes. Kwon et al.  propose PagedAttention~\cite{vllmsosp}, which allows more requests to concurrently execute, eliminating fragmentation in \textit{KV-cache}. The use of Multi Query Attention (MQA) \cite{multiqueryattention}, Group Query Attention (GQA) \cite{groupedqueryattention} in leading edge LLM models like \llama{}2 \cite{touvron2023llama}, Falcon \cite{almazrouei2023falcon} and Yi \cite{yi} has also significantly helped in alleviating memory bottleneck in LLM inference. For instance, \llamaL model has a 8\myx smaller KV-cache footprint compared to \llama-65B.

\algrenewcommand\algorithmicindent{1em}
\begin{algorithm}[t!]
\caption{\Rls. New requests are admitted only if there are no decodes left (line 3). This optimizes TBT but wastes GPU compute in many decode-only iterations (line 10) with potentially small batch sizes.}
\label{alg:request_level}
\begin{algorithmic}[1]
\State Initialize current batch $B \leftarrow \emptyset$
\While{True}
    \If{$B = \emptyset$} %
        \State $R_{new} \leftarrow$ get\_next\_request()
       \While{can\_allocate\_request($R_{new}$)}
            \State $B \leftarrow B + R_{new}$
            \State $R_{new} \leftarrow$ get\_next\_request()
        \EndWhile
        \State prefill($B$)
    \Else
        \State decode($B$)
        \State $B \leftarrow$ filter\_finished\_requests($B$)
    \EndIf
\EndWhile
\end{algorithmic}
\end{algorithm}

\algrenewcommand\algorithmicindent{1em}
\begin{algorithm}[t!]
\caption{\Ils (vLLM). Prefills are executed eagerly (lines 8-9), potentially introducing a generation stall for ongoing decodes (line 12).}
\label{alg:iteration_level}
\begin{algorithmic}[1]
\State Initialize current batch $B \leftarrow \emptyset$
\While{True}
    \State $B_{new} \leftarrow \emptyset$
    \State $R_{new} \leftarrow$ get\_next\_request()
    \While{can\_allocate\_request($R_{new}$)}
        \State $B_{new} \leftarrow B_{new} + R_{new}$ 
        \State $R_{new} \leftarrow$ get\_next\_request()
    \EndWhile
    \If{$B_{new} \neq \emptyset$}
        \State prefill($B_{new}$) %
        \State $B \leftarrow B + B_{new}$
    \Else
        \State decode($B$) %
    \EndIf
    \State $B \leftarrow$ filter\_finished\_requests($B$)
\EndWhile
\end{algorithmic}
\end{algorithm}

\subsection{Multi-GPU LLM Inference}
\label{sec:background_multi_GPU_inference}
With ever-increasing growth in model sizes, it becomes necessary to scale LLMs to multi-GPU or even multi-node deployments~\cite{efficiently-scaling-transformer-inference,multinodeinferenceblog}. Furthermore, LLM inference throughput, specifically that of the decode phase is limited by the maximum batch size we can fit on a GPU. Inference efficiency can therefore benefit from model-parallelism which allows larger batch sizes by sharding model weights across multiple GPUs. Prior work has employed both tensor-parallelism (TP)~\cite{megatron} and pipeline-parallelism (PP)~\cite{orca, fastertransformer, fastserve} for this purpose.

TP shards each layer across the participating GPUs by splitting the model weights and KV-cache equally across GPU workers. This way, TP can linearly scale per-GPU batch size. However, TP involves a high communication cost due to two all-reduce operations per layer -- one in attention computation and the other in FFN~\cite{megatron}. Moreover, since these communication operations are in the critical path, TP is preferred only within a single node where GPUs are connected via high bandwidth interconnects like NVLink.

Compared to TP, PP splits a model layer-wise, where each GPU is responsible for a subset of layers. To keep all GPUs in the `pipeline' busy, \textit{micro-batching} is employed. These micro-batches move along the pipeline from one stage to the next at each iteration. PP has much better compute-communication ratio compared to TP, as it only needs to send activations once for multiple layers of compute. Furthermore, PP requires communication only via point-to-point communication operations, compared to the more expensive allreduces in TP. Thus, PP is more efficient than TP when high-bandwidth interconnects are unavailable \eg in cross-node deployments.

\subsection{Performance Metrics}
\label{subsec:perf_metrics}
There are two primary latency metrics of interest for LLM serving: TTFT (time-to-first-token) and TBT (time-between-tokens). For a given request, TTFT measures the latency of generating the first output token from the moment a request arrives in the system. This metric reflects the initial responsiveness of the model. TBT on the other hand measures the interval between the generation of consecutive output tokens of a request, and affects the overall perceived fluidity of the response. When system is under load, low throughput can lead to large scheduling delays and consequently higher TTFT. 

In addition, we use a throughput metric, \textit{Capacity}, defined as the maximum request load (queries-per-second) a system can sustain while meeting certain latency targets. Higher capacity is desirable because it reduces the cost of serving.

\subsection{Scheduling Policies for LLM Inference}
The scheduler is responsible for admission control and batching policy. For the ease of exposition, we investigate existing LLM inference schedulers by  broadly classifying them under two categories -- \prefillprio and \decodeprio.

Conventional inference engines like FasterTransformer~\cite{fastertransformer}, Triton Inference Server~\cite{triton-batch} use \decodeprio schedules with \rls \ie they pick a batch of requests and execute it until \textit{all} requests in the batch complete (\autoref{alg:request_level}). 
This approach reduces the operational complexity of the scheduling framework but at the expense of inefficient resource utilization. Different requests in a batch typically have a large variation in the number of input and output tokens. Request-level schedulers pad shorter requests with zeros to match their length with the longest request in the batch which results in wasteful compute and longer wait times for pending requests~\cite{orca}.

To avoid wasted compute of \rls, Orca~\cite{orca} introduced a fine-grained \ils mechanism where requests can dynamically enter and exit a batch after each model iteration.(\autoref{alg:iteration_level}). 
This approach can significantly increase system throughput and is being used in many LLM inference serving systems today \eg vLLM \cite{vLLM:github}, TensorRT-LLM \cite{tensorrtllm:github}, and LightLLM \cite{lightllm:github}.

Current \ils systems such as vLLM \cite{vLLM:github} and Orca \cite{orca} use \prefillprio schedules that eagerly admit new requests in a running batch at the first available opportunity, e.g., whenever GPU memory becomes available. Prioritizing prefills can improve throughput because it increases the batch size of subsequent decode iterations.

\section{Motivation}
\label{sec:motivation}
\begin{figure}[t]
    \centering
        \includegraphics[width=0.47\textwidth]{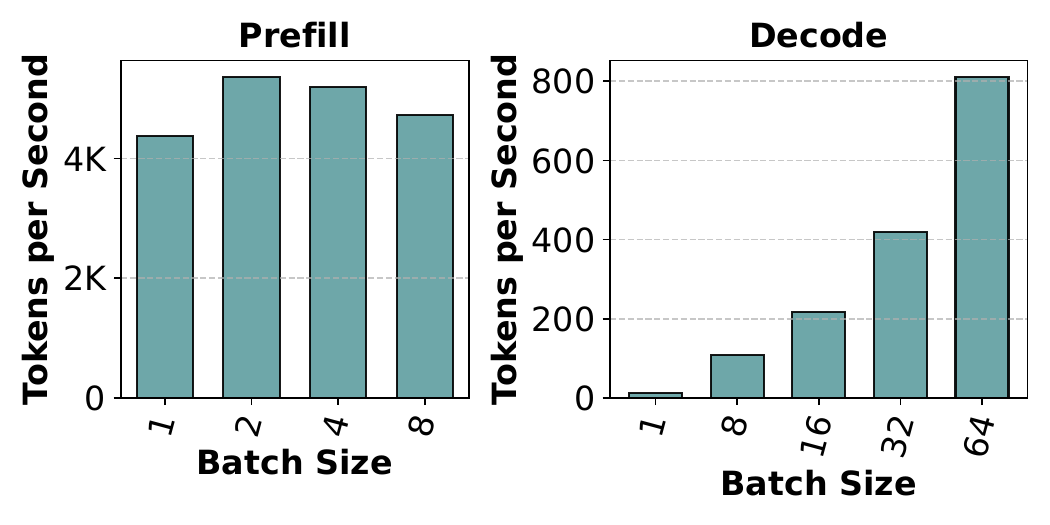}
    \caption{Throughput of the prefill and decode phases with different batch sizes for \mistral running on a single A100 GPU. We use prompt length of 1024 for both prefill and decode experiments. Note that different y-axis, showing prefills are much more efficient than decode. Further, note that \textit{batching boosts decode throughput almost linearly but has a marginal effect on prefill throughput.}}
   \label{fig:background:tput}
\end{figure}

In this section, we first analyse the cost of prefill and decode operations. We then highlight the throughput-latency trade-off and pipeline bubbles that appear in serving LLMs.

\subsection{Cost Analysis of Prefill and Decode}
\label{sec:analysis:throughput}

As discussed in~\autoref{sec:background:inference-process}, while the \textit{prefill} phase processes all input tokens in parallel and effectively saturates GPU compute, the \textit{decode} phase processes only a single token at a time and is very inefficient. \autoref{fig:background:tput} illustrates throughput as a function of batch size, and we can observe that while for decode iterations throughput increases roughly linearly with batch size, prefill throughput almost saturates even with a single request.

\noindent{\bf Takeaway-1:} \textit{The two phases of LLM inference -- prefill and decode -- demonstrate contrasting behaviors wherein batching boosts decode phase throughput immensely but has little effect on prefill throughput.}

\begin{figure}[t]
    \centering
    \includegraphics[width=0.47\textwidth]{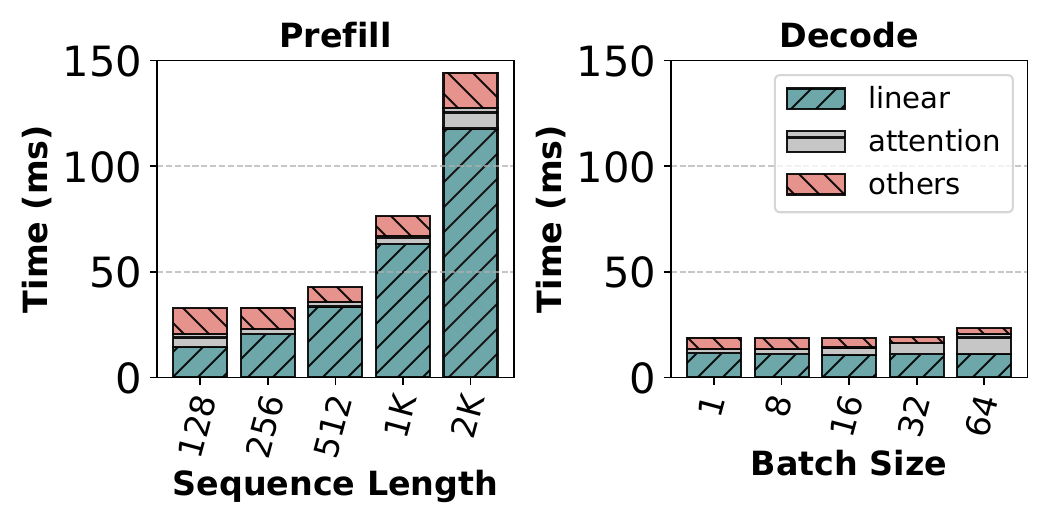}
    \caption{Prefill and decode time with different input sizes for \mistral running on single A100 GPU. Linear layers contribute to the majority of runtime in both prefill and decode phases. Due to the low arithmetic intensity in decode batches, the cost of linear operation for 1 decode token is nearly same as 128 prefill tokens. }
    \label{fig:motivation:op-wise-time}
\end{figure}

\if
To illustrate this, we compute the per-token prefill and decode times in \autoref{fig:analysis_per_token} for different input sizes. We vary the model's input size either by varying prompt length using batch size of one (for prefill) or varying batch size (for decode). We can conclude the following from this experiment.

\noindent{\bf Takeaway-1:} \textit{A single prefill request with modest sequence length can effectively saturate GPU compute}. As can be seen in ~\autoref{fig:analysis_per_token}, the prefill throughput increases almost linearly at low sequence lengths, and saturates quickly around sequence length as low as 512 tokens. Thus, even small sequence lengths can provide enough parallelism in the prefill phase to use the GPU compute effectively. This motivates the use of \chunking in \sysname.

Note that for very long sequence lengths \eg 16K, per-token prefill time increases slightly. This is due to the cost of attention operator which grows quadratically with sequence length.

The saturation point depends the model's hidden size and deployment configuration. For example, models with higher embedding size will saturate prefill throughput at smaller sequence lengths while models deployed with large degree of tensor-parallelism will require more tokens to saturate since the amount of computation per-GPU becomes smaller.

\begin{figure}[t]
    \centering
    \includegraphics[width=0.47\textwidth]{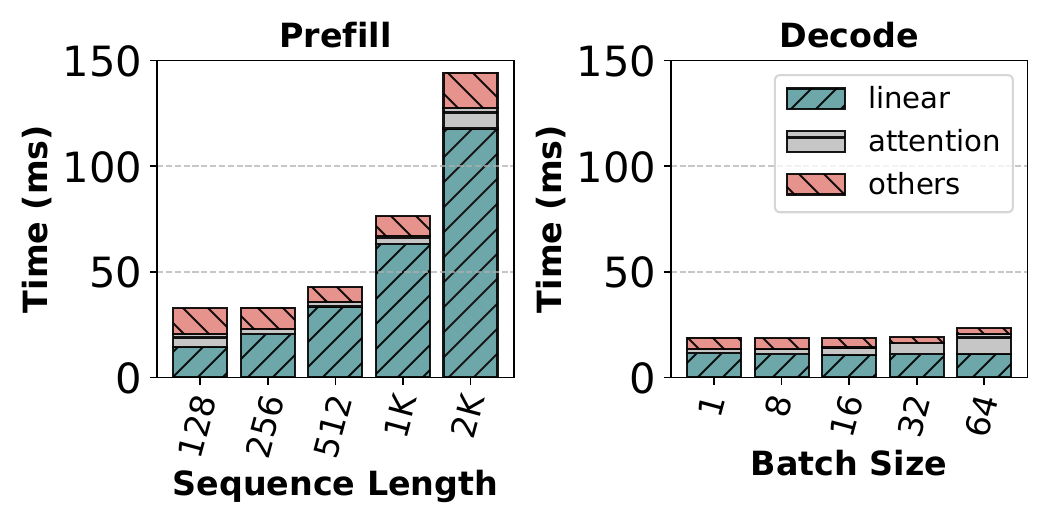}
    \caption{Prefill and decode time with different input sizes for \mistral running on single A100 GPU. Linear layers contribute to the majority of runtime in both prefill and decode phases. Due to the low arithmetic intensity in decode batches, the cost of linear operation for 1 decode token is nearly same as 128 prefill tokens. }
    \label{fig:motivation:op-wise-time}
\end{figure}

\noindent{\bf Takeaway-2:} \textit{The decode time per token is order(s) of magnitude higher than prefill, especially at small batch sizes}. As shown in ~\autoref{fig:analysis_per_token}, at a very high decode batch size of 128, the per-token decode time is still $4.4\times$ higher than the lowest prefill time per-token. Note that the decode time per token improves almost linearly with batch size. At 16, which is in lower side of decode batch sizes, the per-token decode time is still 20X. Thus, optimizing decodes is crucial for efficient LLM inference.
\fi

\autoref{fig:motivation:op-wise-time} breaks down the prefill and decode compute times into linear, attention and others, and shows their individual contributions. From the figure, we see that linear operators contribute to the majority of the runtime cost.  While attention cost grows quadratically with sequence length, linear operators still contribute more than 80\% to the total time even at high sequence lengths. Therefore, optimizing linear operators is important for improving LLM inference.

\noindent{\bf Low Compute Utilization during Decodes:} Low compute utilization during the decode phase is a waste of GPU's processing capacity. To understand this further, we analyze the arithmetic intensity of prefill and decode iterations. Since the majority of the time in LLM inference is spent in linear operators, we focus our analysis on them.

\begin{figure}[t!]
    \centering
        \includegraphics[width=0.45\textwidth]{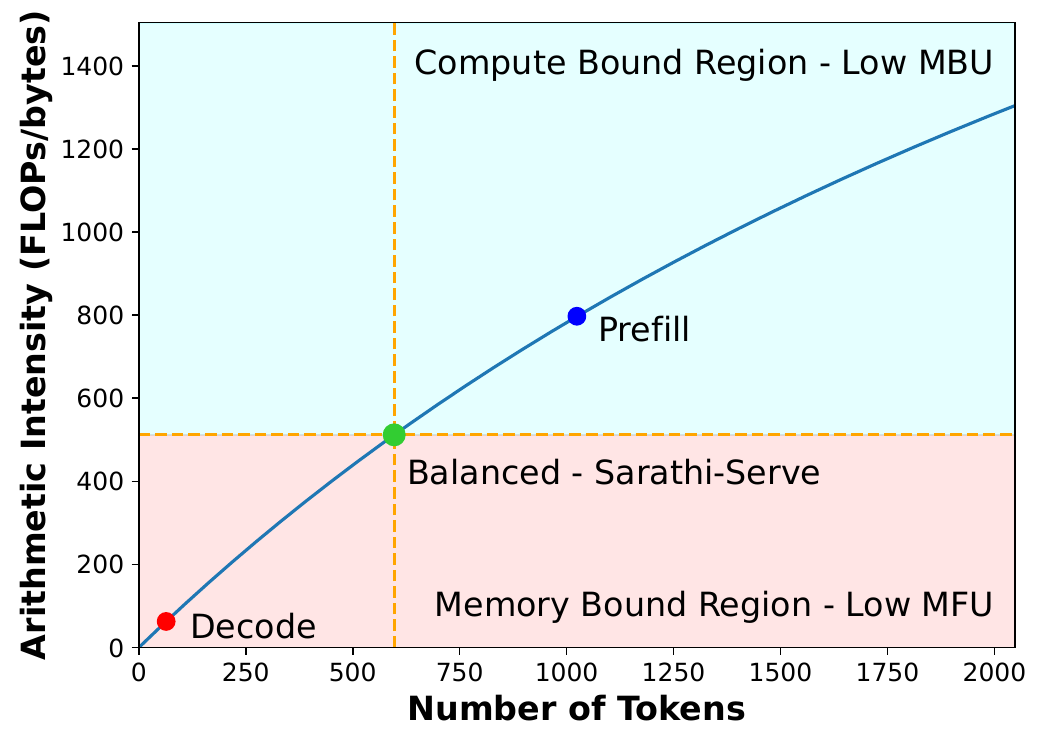}
        
    \caption{Arithmetic intensity trend for \llamaL linear operations with different number of token running on four A100s. Decode batches have low arithmetic intensity \ie they are bottlenecked by memory fetch time, leading to low compute utilization. Prefill batches are compute bound with sub-optimal bandwidth utilization. \sysname forms balanced batches by combining decodes and prefill chunks to maximize both compute and bandwidth utilization.}
       \label{fig:motivation:arithmatic-intensity}
\end{figure}

\begin{figure}[t!]
    \centering
    \hspace{-12px}
    \includegraphics[width=0.45\textwidth]{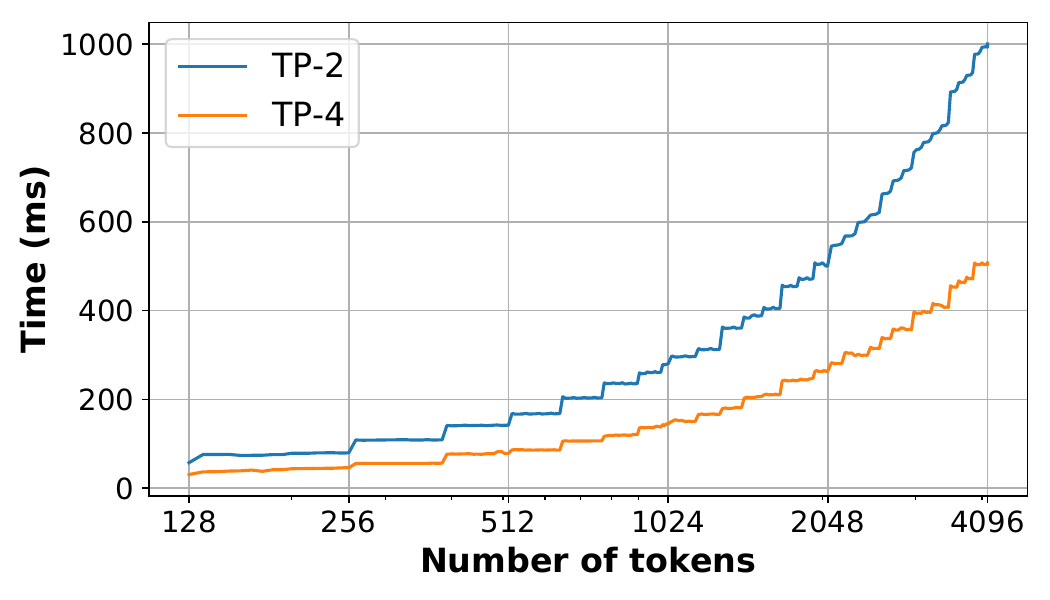}
    \caption{Linear layer execution time as function of number of tokens in a batch for \llamaL on A100(s) with different tensor parallel degrees. When the number of tokens is small, execution time is dictated by the cost of fetching weights from HBM memory. Hence, execution time is largely stagnant in the 128-512 tokens range, especially for higher tensor parallel degrees. Once the number of tokens in the batch cross a critical threshold, the operation become compute bound and the runtime increases linearly with number of tokens.}%
    \label{fig:motivation:mlp-time} 
\end{figure}

Matrix multiplication kernels overlap memory accesses along with computation of math operations. The total execution time of an operation can be approximated to $T = \text{max}(T_{\text{math}}, T_{\text{mem}})$, where $T_{\text{math}}$ and $T_{\text{mem}}$ represent the time spent on math and memory fetch operations respectively. An operation is considered memory-bound if $T_{\text{math}} < T_{\text{mem}}$. Memory-bound operations have low Model FLOPs Utilization (MFU) \cite{chowdhery2022Palm}. On the other hand, compute-bound operations have low Model Bandwidth Utilization (MBU).
When $T_{\text{math}} = T_{\text{mem}}$, both compute and memory bandwidth utilization are maximized. Arithmetic intensity quantifies the number of math operations performed per byte of data fetched from the memory. At the optimal point, the arithmetic  intensity of operation matches the FLOPS-to-Bandwidth ratio of the device. \autoref{fig:motivation:arithmatic-intensity} shows arithmetic intensity as a function of the number of tokens in the batch for linear layers in \llamaL running on four A100 GPUs. Prefill batches amortize the cost of fetching weights of the linear operators from HBM memory to GPU cache over a large number of tokens, allowing it to have high arithmetic intensity. In contrast, decode batches have very low computation intensity. \autoref{fig:motivation:mlp-time} shows the total execution time of linear operators in an iteration for \llamaL as a function of the number of tokens. Note that execution time increases only marginally in the beginning \ie as long as the batch is in a memory-bound regime, but linearly afterwards \ie when the batch becomes compute-bound.\footnote{Theoretically, we expect the operators to become compute-bound at $\sim$200 tokens on A100 GPUs, however, in practice we observe that it happens at $\sim$500-600 tokens for higher tensor parallel dimensions due to fixed overheads.}

\noindent{\bf Takeaway-2:} \textit{Decode batches operate in memory-bound regime leaving compute underutilized. This implies that more tokens can be processed along with a decode batch without significantly increasing its latency.} %

\begin{figure}[t!]
    \centering
        \includegraphics[width=0.47\textwidth]{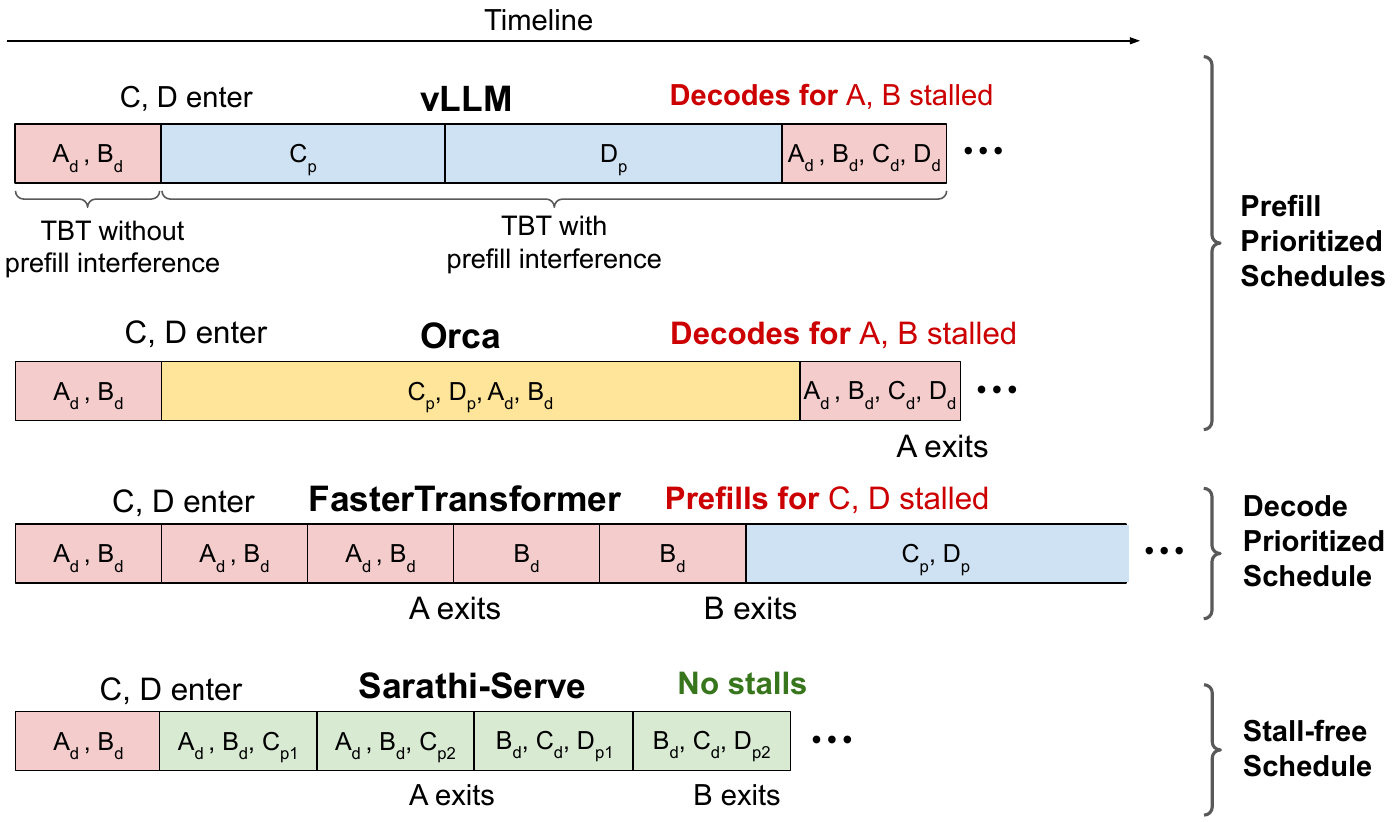}
    \caption{A generation stall occurs when one or more prefills are scheduled in between consecutive decode iterations of a request. A, B, C and D represent different requests. Subscript $d$ represents a decode iteration, $p$ represents a full prefill and $p0$, $p1$ represent two chunked prefills of a given prompt. vLLM induces generation stalls by scheduling as many prefills as possible before resuming ongoing decodes. Despite supporting hybrid batches, Orca cannot mitigate generation stalls because the execution time of batches containing long prompts remains high. FasterTransformer is free of generation stalls as it finishes all ongoing decodes before scheduling a new prefill but compromises on throughput due to low decode batch size. In contrast, \sysname generates a schedule that eliminates generation stalls yet delivers high throughput.}
        \label{fig:motivation:generationstalls}
\end{figure}

\subsection{Throughput-Latency Trade-off}
\label{sec:latencyandstalls}

\Ils improves system throughput but we show that it comes at the cost of high TBT latency due to a phenomenon we call \textit{generation stalls}.

\autoref{fig:motivation:generationstalls} compares different scheduling policies. The example shows a timeline (left to right) of requests A, B, C and D. Requests A and B are in decode phase at the start of the interval and after one iteration, requests C and D enter the system. Orca and vLLM both use FCFS \ils with eager admission of prefill requests %
but differ in their batch composition policy. Orca supports hybrid batches composed of both prefill and decode requests whereas vLLM only supports batches that contain either all prefill or all decode requests. Irrespective of this difference, both Orca and vLLM can improve throughput by maximizing the batch size in subsequent decode iterations. However, eagerly scheduling prefills of requests C and D delays the decodes of already running requests A and B because an iteration that computes one or more prefills can take several seconds depending on the lengths of input prompts. Therefore, \prefillprio schedulers can introduce \textit{generation stalls} for ongoing decodes resulting in latency spikes caused by high TBT.

In contrast to \ils, \rls systems such as FasterTransformer \cite{fastertransformer} do not schedule new requests until \textit{all} the already running requests complete their decode phase (line 3 in \autoref{alg:request_level}). In \autoref{fig:motivation:generationstalls}, the prefills for requests C and D get stalled until requests A and B both exit the system. Therefore, \decodeprio systems provide low TBT latency albeit at the cost of low system throughput. For example, Kwon et al. \cite{vllmsosp} show that \ils with PagedAttention can achieve an order of magnitude higher throughput compared to FasterTransformer.

One way to reduce latency spikes in \ils systems is to use smaller batch sizes as recommended in Orca~\cite{orca}. However, lowering batch size adversely impacts throughput as shown in~\autoref{sec:background:inference-process}. Therefore, existing systems are forced to trade-off between throughput and latency depending on the desired SLOs.

\noindent{\bf Takeaway-3:} \textit{The interleaving of prefills and decodes involves a trade-off between throughput and latency for current LLM inference schedulers. State-of-the-art systems today use prefill-prioritizing schedules that trade TBT latency for high throughput.}

\begin{figure}[!t]
    \centering
    \includegraphics[width=0.43\textwidth]{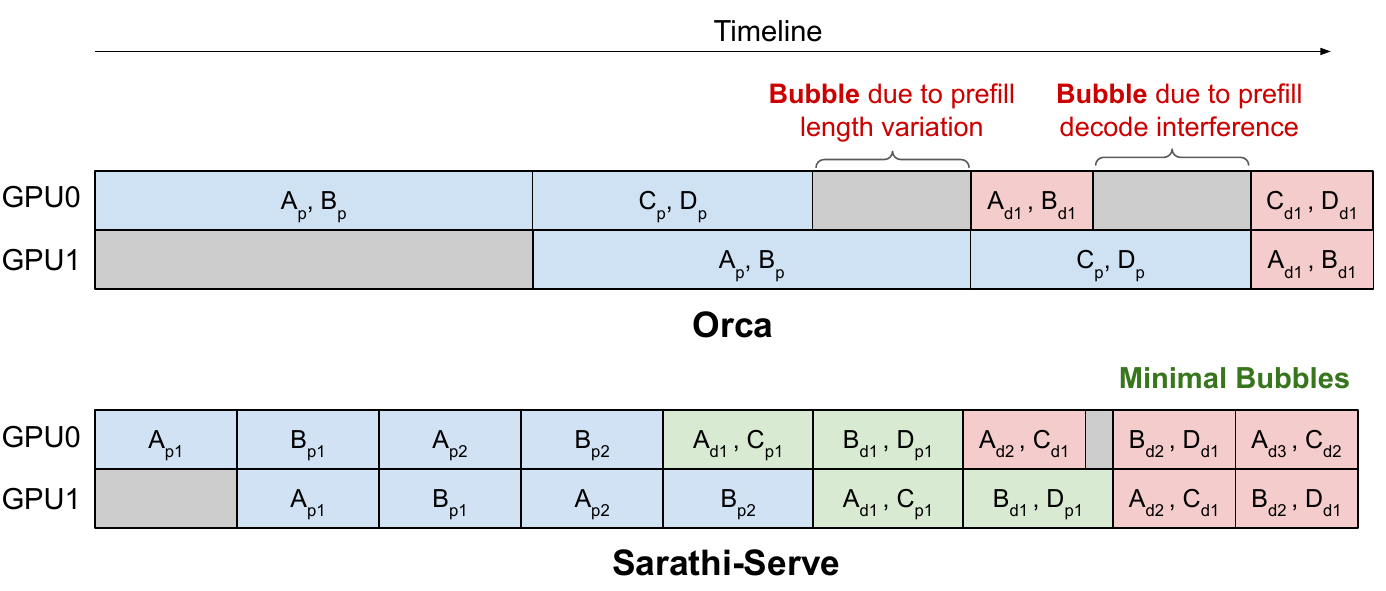}
    \caption{A 2-way pipeline parallel iteration-level schedule in Orca across 4 requests (A,B,C,D) shows the existence of pipeline bubbles due to non-uniform batch execution times. \sysname is able to minimize these stalls by creating uniform-compute batches.}
    \label{fig-mot-bubbles}
\end{figure}

\subsection{Pipeline Bubbles waste GPU Cycles}
Pipeline-parallelism (PP) is a popular strategy for cross-node deployment of large models, owing to its lower communication overheads compared to Tensor Parallelism (TP). A challenge with PP, however, is that it introduces \textit{pipeline bubbles} or periods of GPU inactivity as subsequent pipeline stages have to wait for the completion of the corresponding micro-batch in the prior stages. Pipeline bubbles is a known problem in training jobs, where they arise between the forward and backward passes due to prior stages needing to wait for the backward pass to arrive. Micro-batching is a common technique used in PP training jobs to mitigate pipeline bubbles~\cite{varuna,pipedream,gpipe}. 

Inference jobs only require forward computation and therefore one might expect that micro-batching can eliminate pipeline bubbles during inference. In fact, prior work on transformer inference, such as, FasterTransformer~\cite{fastertransformer} and FastServe~\cite{fastserve} use micro-batches but do not mention pipeline-bubbles.  Recently proposed Orca~\cite{orca} also suggests that iteration-level scheduling eliminates bubbles in pipeline scheduling (see Figure 8 in ~\cite{orca}). However, our experiments show that even with iteration-level scheduling, pipeline bubbles can waste significant GPU cycles with PP (\autoref{sec-eval-pp}).

Each micro-batch (or iteration) in LLM inference can require a different amount of compute (and consequently has varying execution time), depending on the composition of prefill and decode tokens in the micro-batch (see~\autoref{fig-mot-bubbles}). We identify three types of bubbles during inference: (1) bubbles like $PB_1$ that occur due to the varying number of prefill tokens in two consecutive micro-batches (2) bubbles like $PB_2$ that occur due to different compute times of prefill and decode stages when one is followed by the other, and (3) bubbles like $PB_3$ that occur due to difference in decode compute times between micro-batches since the attention cost depends on the accumulated context length  (size of the KV-cache) and varies across requests. For Falcon-180B, a single prompt of 4k tokens takes $\approx1150$ ms to execute compared to a decode only iteration with batch size 32 which would take about $\approx200$ ms to execute. Interleaving of these iteration could result in a bubble of $\approx950$ ms. These pipeline bubbles are wasted GPU cycles and directly correspond to a loss in serving throughput and increased latency. This problem is aggravated with increase in prompt lengths and batch size, due to longer and more frequent prefill iterations respectively. If we can ensure that each micro-batch performs uniform computation, we can mitigate these pipeline bubbles.

\noindent{\bf Takeaway-4:} \textit{There can be a large variance in compute time of LLM iterations depending on composition of prefill- and decode-tokens in the batch. This can lead to significant bubbles when using pipeline-parallelism.}

\section{\sysname : Design and Implementation}\label{sec:design}
 We now discuss the design and implementation of \sysname{} --- a system that provides high throughput with predictable tail latency via two key techniques -- \chunking and \hbatch.

\begin{figure*}[t!]
    \centering
        \begin{subfigure}[b]{\linewidth}
        \includegraphics[width=\linewidth]{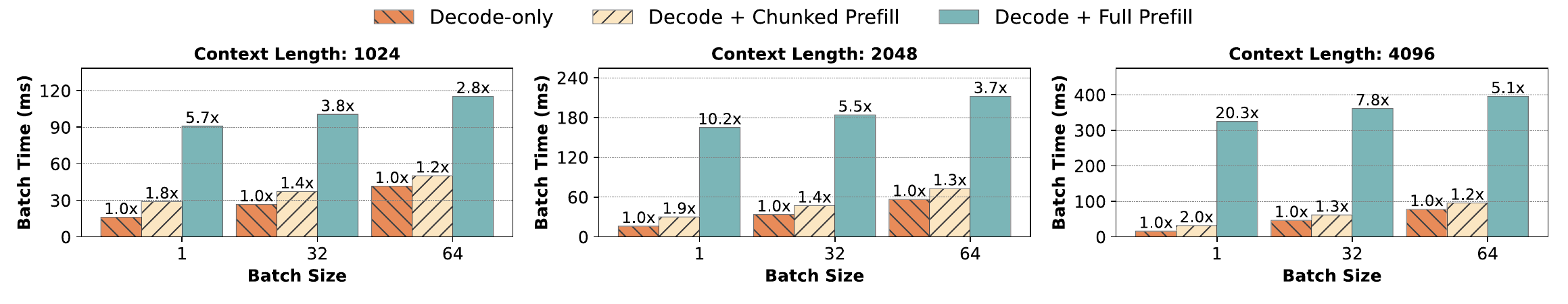}
        \caption{Mistral-7B on one A100s with token budget of 256.}
        \label{fig:motivation:incrementalcost:mistral}
    \end{subfigure}
    \\
    \begin{subfigure}[b]{\linewidth}
        \includegraphics[width=\linewidth]{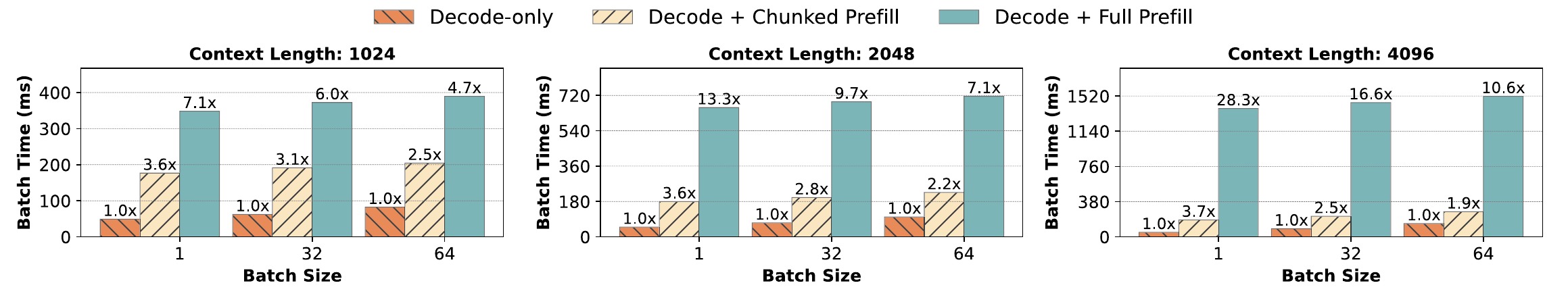}
        \caption{\llamaL on four A100s with token budget of 512.}
        \label{fig:motivation:incrementalcost:llama}
    \end{subfigure}

    \caption{The incremental cost of coalescing prefills with decode batches. We consider two batching schemes -- (i) Decode + Full Prefill represents the hybrid batching of Orca wherein the entire prefill is executed in a single iteration along with ongoing decodes. (ii) Decode + Chunked Prefill represents \sysname wherein prefills are chunked before being coalesced with ongoing decodes with a fixed token budget. \sysname processes prefill tokens with much lower impact on the latency of decodes. Further, the relative impact of \sysname on latency reduces with higher decode batch size and context lengths.}
        \label{fig:motivation:incrementalcost}
\end{figure*}

\subsection{Chunked-prefills}
\label{sec-design-chunkedprefill}

As we show in \autoref{sec:analysis:throughput}, decode batches are heavily memory bound with low arithmetic intensity. This slack in arithmetic intensity presents an opportunity to piggyback additional computation in decode batches. Naively, this can be done by creating hybrid batches which combine the memory bound decodes along with compute bound prefills. However, in many practical scenarios, input prompts contain several thousand tokens on average \eg \autoref{table:eval:datasets} shows that the median prompt size in \sharegpt and \arxivsum datasets is 1730 and 7059 respectively. Combining these long prefills with decode iterations would lead to high TBT latency.

To tackle this challenge, we present a technique called \chunking which allows computing large prefills in small chunks across several iterations. \Chunking is a prefill splitting mechanism hinged on two key insights. First, as discussed in~\autoref{sec:analysis:throughput}, a prefill request with modest sequence length can effectively saturate GPU compute. For example, in~\autoref{fig:motivation:op-wise-time}, prefill throughput starts saturating around sequence length of 512 tokens. Second, in many practical scenarios, input prompts contain several thousand tokens on average (\autoref{table:eval:datasets}). This provides an opportunity to break large prefill requests into smaller units of compute which are still large enough to saturate GPU compute. In \sysname, we leverage this mechanism to form batches with appropriate number of tokens such that we can utilize the compute potential in decode batches without violating the TBT SLO.

\algrenewcommand\algorithmicindent{1em}
\begin{algorithm}[t!]
\caption{\Hbatch with \sysname.  First the batch is filled with with ongoing decode tokens (lines 6-8) and optionally one prefill chunk from ongoing (lines 10-12). Finally, new requests are added (lines 13-20) within the token budget so as to maximize throughput with minimal latency impact on the TBT of delaying the ongoing decodes.}
\label{alg:stall_free}
\begin{algorithmic}[1]
\State \textbf{Input:} $T_{max}$, Application TBT SLO.
\State Initialize \textit{token\_budget}, $\tau$ $\leftarrow$ compute\_token\_buget($T_{max}$)
\State Initialize \textit{batch\_num\_tokens}, $n_t$ $\leftarrow$ 0
\State Initialize current batch $B \leftarrow \emptyset$

\While{True}
    \For{$R$ in $B$}
        \If{is\_prefill\_complete($R$)} %
            \State $n_t \leftarrow n_t + 1$
        \EndIf
    \EndFor
    
    \For{$R$ in $B$}
        \If{not is\_prefill\_complete($R$)} %
            \State $c \leftarrow$  get\_next\_chunk\_size($R$, $\tau$, $n_t$)
            \State $n_t \leftarrow n_t + c$
        \EndIf
    \EndFor
    
    \State $R_{new} \leftarrow$ get\_next\_request()

    \While {can\_allocate\_request($R_{new}$)  $ \land$ $n_t$ < $\tau$} %
        \State $ c \leftarrow$  get\_next\_chunk\_size($R_{new}$, $\tau$, $n_t$)
        
        \If{$c$ > 0}
            \State $n_t \leftarrow n_t + c$
            \State $B \leftarrow R_{new}$ 
        \Else
            \State break
        \EndIf
    \EndWhile

    \State
    \State process\_hybrid\_batch($B$)
    \State $B \leftarrow$ filter\_finished\_requests($B$)
    \State $n_t \leftarrow 0$
\EndWhile
\end{algorithmic}
\end{algorithm}

\subsection{\Hbatch}

The \sysname scheduler is an iteration-level scheduler that leverages \chunking and coalescing of prefills and decodes to improve throughput while minimizing latency.

Unlike Orca and vLLM which stall existing decodes to execute prefills, \sysname leverages the arithmetic intensity slack in decode iterations to execute prefills without delaying the execution of decode requests in the system. We call this approach \hbatch (\autoref{alg:stall_free}). \sysname first calculates the budget of maximum number of tokens that can be executed in a batch based on user specified SLO. We describe the considerations involved in determining this token budget in depth in \autoref{section:howtopickachunksize}. In every scheduling iteration, we first pack all the running decodes in the next batch (lines 6-8 in \autoref{alg:stall_free}). After that, we include any partially completed prefill (lines 9-12). Only after all the running requests have been accommodated, we admit new requests (lines 13-20). When adding prefill requests to the batch, we compute the maximum chunk size that can be accommodated within the leftover token budget for that batch (lines 11, 15). By restricting the computational load in every iteration, \hbatch ensures that decodes never experience a generation stall due to a co-running prefill chunk.  We compare the latency for hybrid batches with and without chunked prefills in \autoref{fig:motivation:incrementalcost}. Naive hybrid batching leads to dramatic increase of up to 28.3\myx in the TBT latency compared to a decode-only batch. In contrast, \sysname provides a much tighter bound on latency with chunking.

\autoref{fig:motivation:generationstalls} shows the scheduling policy of \sysname in action, for the same example used in~\autoref{sec:latencyandstalls}. The first iteration is decode-only as there are no prefills to be computed. However, after a new request C enters the system, \sysname first splits the prefill of C into two chunks and schedules them in subsequent iterations. At the same time, with \hbatch, it coalesces the chunked prefills with ongoing decodes of A and B. This way, \sysname stalls neither decodes nor prefills unlike existing systems, allowing \sysname to be largely free of latency spikes in TBT without compromising throughput. Furthermore, \hbatch combined with \chunking also ensures uniform compute hybrid batches in most cases, which helps reduce bubbles when using pipeline parallelism, thereby enabling efficient and scalable deployments.

\subsection{Determining Token Budget}
\label{section:howtopickachunksize}

The token budget is determined based on two competing factors --- TBT SLO requirement and \chunking overhead. From a TBT minimization point of view, a smaller token budget is preferable because iterations with fewer prefill tokens have lower latency. However, smaller token budget can result in excessive chunking of prefills resulting in overheads due to 1) lower GPU utilization and 2) repeated KV-cache access in the attention operation which we discuss below.

During the computation of \chunking, the attention operation for every chunk of a prompt needs to access the KV-cache of \textit{all} prior chunks of the same prompt. This results in increased memory reads from the GPU HBM even though the computational cost is unchanged. For example, if a prefill sequence is split into $N$ chunks, then the first chunk's KV-cache is loaded $N-1$ times, the second chunk's KV-cache is loaded $N-2$ times, and so on. However, we find that even at small chunk sizes attention prefill operation is compute bound operation. In practice, there can be small overhead associated with chunking due to fixed overheads of kernel launch, etc. We present a detailed study of the overheads of \chunking in \autoref{sec-eval-ablation}.

Thus, one needs to take into account the trade-offs between prefill overhead and decode latency while determining the token budget. This can be handled with a one-time profiling of batches with different number of tokens and setting the token budget to maximum number of tokens that can be packed in a batch without violating TBT SLO. 

Another factor that influences the choice of token budget is the \textit{tile-quantization} effect~\cite{tile-quantization}. GPUs compute matmuls by partitioning the given matrices into tiles and assigning them to different thread blocks for parallel computation. Here, each thread block refers to a group of GPU threads and computes the same number of arithmetic operations. Therefore, matmuls achieve maximum GPU utilization when the matrix dimensions are divisible by the tile size.  Otherwise, due to \textit{tile-quantization}, some thread blocks perform extraneous computation~\cite{tile-quantization}. We observe that tile-quantization can dramatically increase prefill computation time \eg in some cases, using chunk size of 257 can increase prefill time by 32\% compared to that with chunk size 256.

Finally, when using pipeline parallelism the effect of token budget on pipeline bubbles should also be taken into account. Larger chunks lead to higher inter-batch runtime variations that result in pipeline bubbles which results in lower overall system throughput. On the other hand, picking a very small token budget can lead to higher overhead due to lower arithmetic intensity and other fixed overheads.

Therefore, selecting a suitable token budget is a complex decision which depends on the desired TBT SLO, parallelism configuration, and specific hardware properties. We leverage Vidur~\cite{vidur}, a LLM inference profiler and simulator to determine the token budget that maximizes system capacity under specific deployment scenario.

\subsection{Implementation} We implement \sysname on top of the open-source implementation of vLLM \cite{vllmsosp, vLLM:github}. %
We added support for paged chunk prefill using FlashAttention v2 \cite{flashattention2} and FlashInfer \cite{flashinfer} kernels. We use FlashAttention backend for all the evaluations in this paper due to its support for wider set of models. We also extend the base vLLM codebase to support various scheduling policies, chunked prefills, pipeline parallelism and an extensive telemetry system. We use NCCL \cite{nccl} for both pipeline and tensor parallel communication.
Source code for the project is available at ~\href{https://github.com/microsoft/sarathi-serve}{https://github.com/microsoft/sarathi-serve}. %

\begin{table}[tb!]
\centering
\setlength{\tabcolsep}{2pt}
\scalebox{0.78}{
\begin{tabular}{r|c|c|c}
\multirow{2}{*}{\textbf{Model}} & \textbf{Attention} & \textbf{GPU} & \textbf{Memory} \\
& \textbf{Mechanism} & \textbf{Configuration} & \textbf{Total (per-GPU)} \\ \toprule
Mistral-7B & GQA-SW & 1 A100 & 80GB (80GB) \\
Yi-34B & GQA & 2 A100s (TP2) & 160GB (80GB)\\
LLaMA2-70B & GQA & 8 A40s (TP4-PP2) & 384GB (48GB) \\
Falcon-180B & GQA & 4 A100s\myx{}2 nodes (TP4-PP2) & 640GB (80GB) \\
\bottomrule
\end{tabular}}
\caption{Models and GPU configurations (GQA: grouped-query attention, SW: sliding window).}
\label{table:eval:models}
\end{table}

\section{Evaluation}
\label{sec-eval}

We evaluate \sysname on a variety of popular models and GPU configurations (see~\autoref{table:eval:models}) and two datasets (see~\autoref{table:eval:datasets}). We consider vLLM and Orca as baseline because they represent the state-of-the-art for LLM inference.  Our evaluation seeks to answer the following questions:
\begin{enumerate}[leftmargin=*] \itemsep0em 
    \item What is the maximum load a model replica can serve under specific Service Level Objective (SLO) constraints with different inference serving systems (\sref{sec-eval-capacity}) and how does this load vary with varying SLO constraints (\sref{subsec-tput-latency})? %
    \item How does \sysname perform under various deployments such as TP and PP? (\sref{sec-eval-pp}) %
    \item What is the overhead of \chunking? 
  (\sref{sec-eval-ablation1})
    \item What is the effect of each of \chunking and \hbatch in isolation as opposed to using them in tandem? (\sref{sec-eval-ablation2})
\end{enumerate}

\begin{table}[t!]
\scalebox{0.78}{
\begin{tabular}{r|ccc|ccc}
\multirow{2}{*}{\textbf{Dataset}} & \multicolumn{3}{c|}{\textbf{Prompt Tokens}} & \multicolumn{3}{c}{\textbf{Output Tokens}}  \\ \cline{2-7}
 &  Median & P90 & Std. & Median & P90 & Std. \\ \toprule
\sharegpt & 1730 & 5696 & 2088 & 415 & 834 & 101 \\
\arxivsum & 7059 & 12985 & 3638 & 208 & 371 & 265\\ \bottomrule
\end{tabular}}
\caption{Datasets used for evaluation.}
\label{table:eval:datasets}
\end{table}

\begin{table}[b!]
\centering
\setlength{\tabcolsep}{2pt}
\scalebox{1}{
\begin{tabular}{r|c|c}
\multirow{2}{*}{\textbf{Model}} & \textbf{\relaxed \slo} & \textbf{\strict \slo} \\
&  \textbf{P99 TBT (s)} & \textbf{P99 TBT (s)} \\ \toprule
Mistral-7B & 0.5 & 0.1 \\
Yi-34B & 1 & 0.2 \\
LLaMA2-70B & 5 & 1 \\
Falcon-180B & 5 & 1 \\
\bottomrule
\end{tabular}}
\caption{\slos for different model configurations.}
\label{table:eval:slos}
\end{table}

\noindent{\bf Models and Environment:} We evaluate \sysname across four different models \mistral \cite{jiang2023mistral}, \yi \cite{yi}, \llamaL \cite{touvron2023llama} and \falcon \cite{almazrouei2023falcon} -- these models are among the best in their model size categories. We use two different server configurations. For all models except \llamaL we use Azure NC96ads v4 VMs, each equipped with 4 NVIDIA 80GB A100 GPUs, connected with pairwise NVLINK. The machines are connected with a 100 Gbps ethernet connection. For \llamaL, we use a server with eight pairwise connected NVIDIA 48GB A40 GPUs. We run \yi in a 2-way tensor parallel configuration (TP-2), and \llamaL and \falcon in a hybrid parallel configuration with four tensor parallel workers and two pipeline stages for (TP4-PP2).

\noindent{\bf Workloads:} In order to emulate the real-world serving scenarios, we generate traces by using the request length characteristics from the \sharegpt \cite{wang2023openchat} and \arxivsum \cite{cohan-etal-2018-discourse} datasets (\autoref{table:eval:datasets}). The \sharegpt trace contains user-shared conversations with ChatGPT-4 \cite{chatgpt}. A conversation may contain multiple rounds of interactions between the user and chatbot. Each such interaction round is performed as a separate request to the system. This multi-round nature leads to high relative variance in the prompt lengths. In contrast, \arxivsum is a collection of scientific publications and their summaries (abstracts) on arXiv.org \cite{arxiv}. This dataset contains longer prompts and lower variance in the  number of output tokens, and is representative of LLM workloads such as Microsoft M365 Copilot \cite{microsoftcopilot} and Google Duet AI \cite{googleduetai} etc. The request arrival times are generated using Poisson distribution. We filter outliers of these datasets by removing requests with total length more than 8192 and 16384 tokens, respectively.

\noindent{\bf Metrics:} We focus on the median value for the TTFT since this metric is obtained only once per user request and on the 99th percentile (P99) for TBT values since every decode token results in a TBT latency value.

\begin{figure}[t!]
    \begin{subfigure}[b]{0.5\textwidth}
        \includegraphics[width=0.95\textwidth]{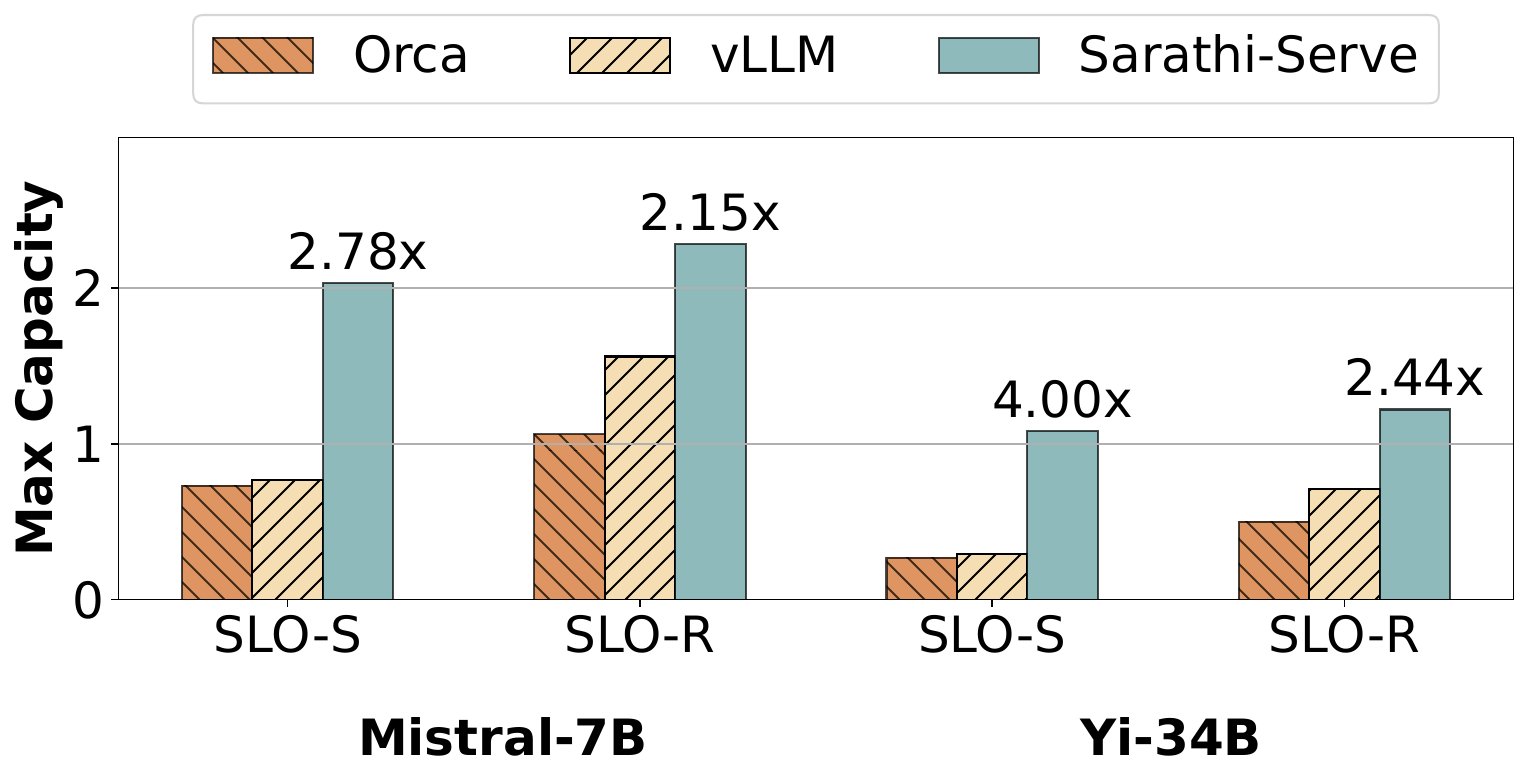}
        \caption{Dataset: \sharegpt.}
        \label{fig:eval:capacity:tp:openchat}
    \end{subfigure}
    \\
    \begin{subfigure}[b]{0.5\textwidth}
        \includegraphics[width=0.95\textwidth]{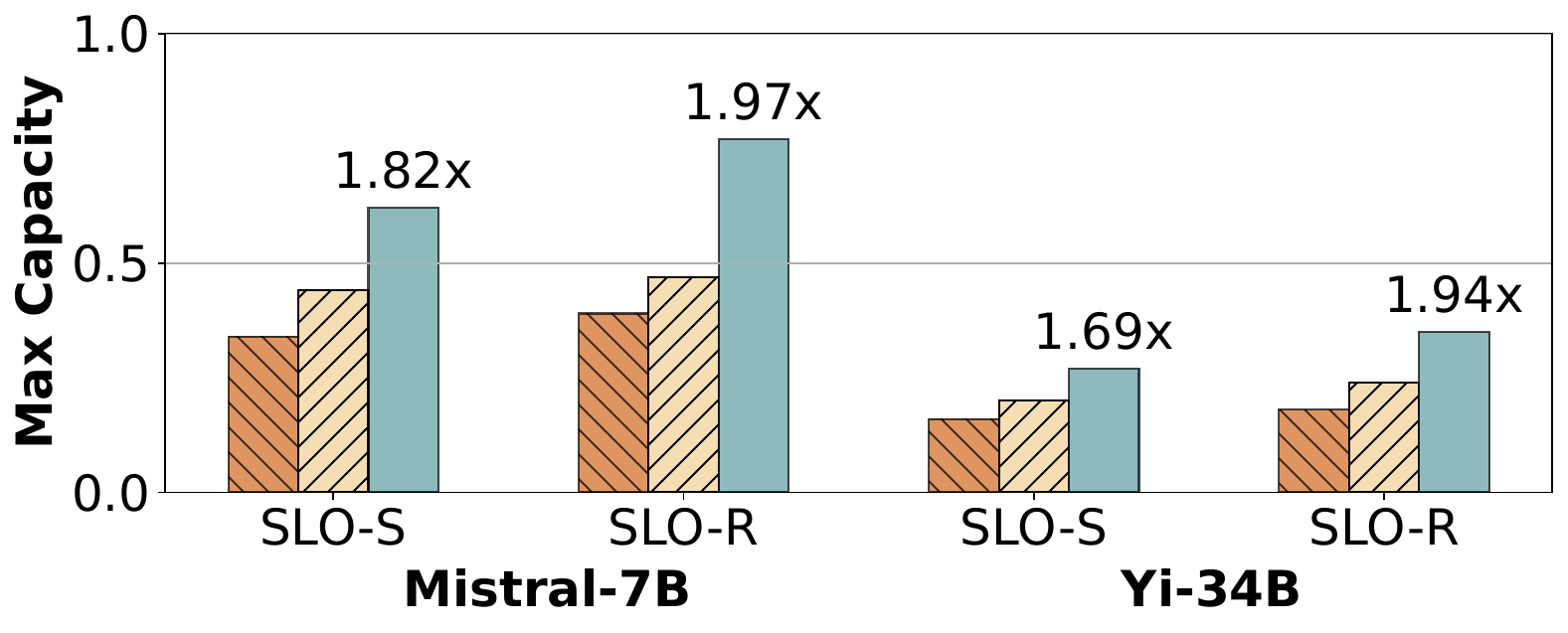}
        \caption{Dataset: \arxivsum.}
        \label{fig:eval:capacity:tp:arxiv}
    \end{subfigure}
    \caption{Capacity (in queries per second) of \mistral and \yi with different schedulers under strict (SLO-S) and relaxed (SLO-R) latency SLOs.}
    \label{fig:eval:capacitytp}
\end{figure}

\subsection{Capacity Evaluation}
\label{sec-eval-capacity}
\begin{figure}[t]
    \begin{subfigure}[b]{0.5\textwidth}
        \includegraphics[width=0.95\textwidth]{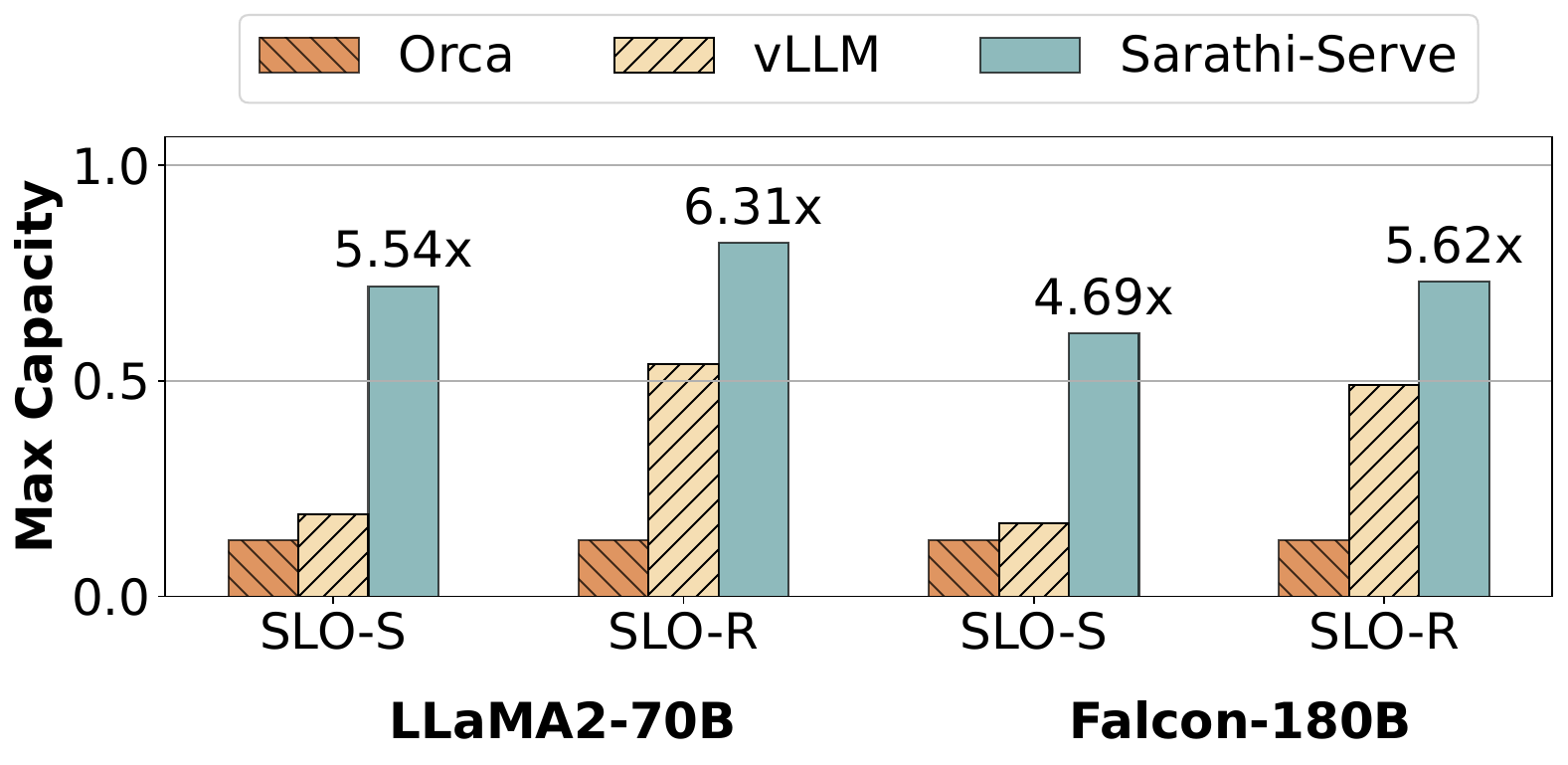}
        \caption{Dataset: \sharegpt.}
        \label{fig:eval:capacity:pp:openchat}
    \end{subfigure}
    \\
    \begin{subfigure}[b]{0.5\textwidth}
        \includegraphics[width=0.95\textwidth]{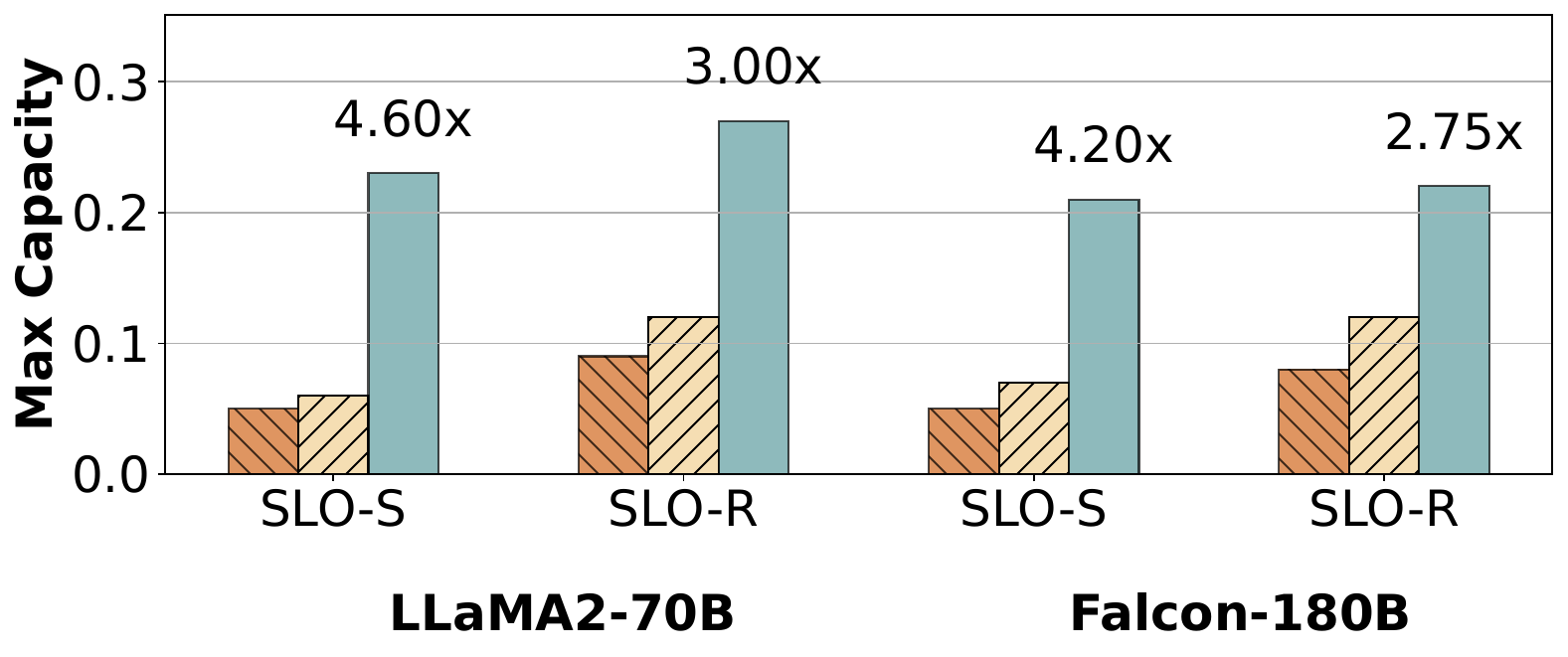}
        \caption{Dataset: \arxivsum.}
        \label{fig:eval:capacity:pp:arxiv}
    \end{subfigure}
    \caption{Capacity of \llamaL and \falcon (models with pipeline parallelism) with different schedulers under strict (SLO-S) and relaxed (SLO-R) latency SLOs.}
    \label{fig:eval:capacitypp}
\end{figure}

We evaluate \sysname, Orca and vLLM on all four models and both datasets under two different latency configurations: \relaxed and \strict. Similar to Patel et al. \cite{patel2023splitwise}, to account for the intrinsic performance limitations of a model and hardware pair, we define the \slo on P99 TBT to be equal to 5\myx and 25\myx the execution time of a decode iteration for a request (with prefill length of 4k and 32 batch size) running without any prefill interference for the \strict and \relaxed settings, respectively. %
~\autoref{table:eval:slos} shows a summary of the absolute \slo thresholds. Note that the \strict SLO represents the latency target desired for interactive applications like chatbots. On the other hand, the \relaxed configuration is exemplary of systems where the complete sequence of output tokens should be generated within a predictable time limit but the TBT constraints on individual tokens is not very strict. For all load experiments, we ensure that the maximum load is sustainable, i.e., the queuing delay does not blow up (we use a limit of 2 seconds on median scheduling delay).

\autoref{fig:eval:capacitytp} and~\autoref{fig:eval:capacitypp} show the results of our capacity experiments. \sysname consistently outperforms Orca and vLLM in all cases across models and workloads. Under \strict \slo, \sysname can sustain up to $4.0\times$ higher load compared to Orca and $3.7\times$ higher load than vLLM under \strict \slo (\yi, \sharegpt). For larger models using pipeline parallelism, \sysname achieves gains of up to 6.3\myx and 4.3\myx compared to Orca and vLLM respectively (\llamaL, \sharegpt) due to few pipeline bubbles.

We observe that in most scenarios, Orca and vLLM violate the P99 TBT latency \slo before they can reach their maximum serviceable throughput. Thus, we observe relaxing the latency target leads to considerable increase in their model serving capacity. In \sysname, one can adjust the chunk size based on the desired SLO. Therefore, we use a strict token budget and split prompts into smaller chunks when operating under \strict latency \slo. This reduces system efficiency marginally but allows us to achieve lower tail latency. On the other hand, when the latency constraint is relaxed, we increase the token budget to allow more efficient prefills. We use token budget of 2048 and 512 for all models under the \relaxed and \strict settings,  respectively, except for the \llamaL \relaxed configuration where we use token budget of 1536 to reduce the impact of pipeline bubbles. The system performance can be further enhanced by dynamically varying the token budget based on workload characteristics. We leave this exploration for future work.

We further notice that vLLM significantly outperforms Orca under relaxed setting. The reason for this is two-fold. First, Orca batches prompts for multiple requests together ({\it max sequence length * batch size} compared to {\it max sequence length} in vLLM), which can lead to even higher tail latency in some cases. Second, vLLM supports a much larger batch size compared to Orca. The lower batch size in Orca is due to the lack of PagedAttention and the large activation memory footprint associated with processing batches with excessively large number of tokens.

Finally, note that the capacity of each system is higher for \sharegpt dataset compared to the \arxivsum dataset. This is expected because prompts in the \arxivsum datasets are much longer - 7059 vs 1730 median tokens as shown in~\autoref{table:eval:datasets}. The larger prompts makes Orca and vLLM more susceptible to latency violations due to higher processing time of these longer prefills.%

 \subsection{Throughput-Latency Tradeoff}
\label{subsec-tput-latency}

\begin{figure}[t!]
\centering
\includegraphics[width=0.95\linewidth]{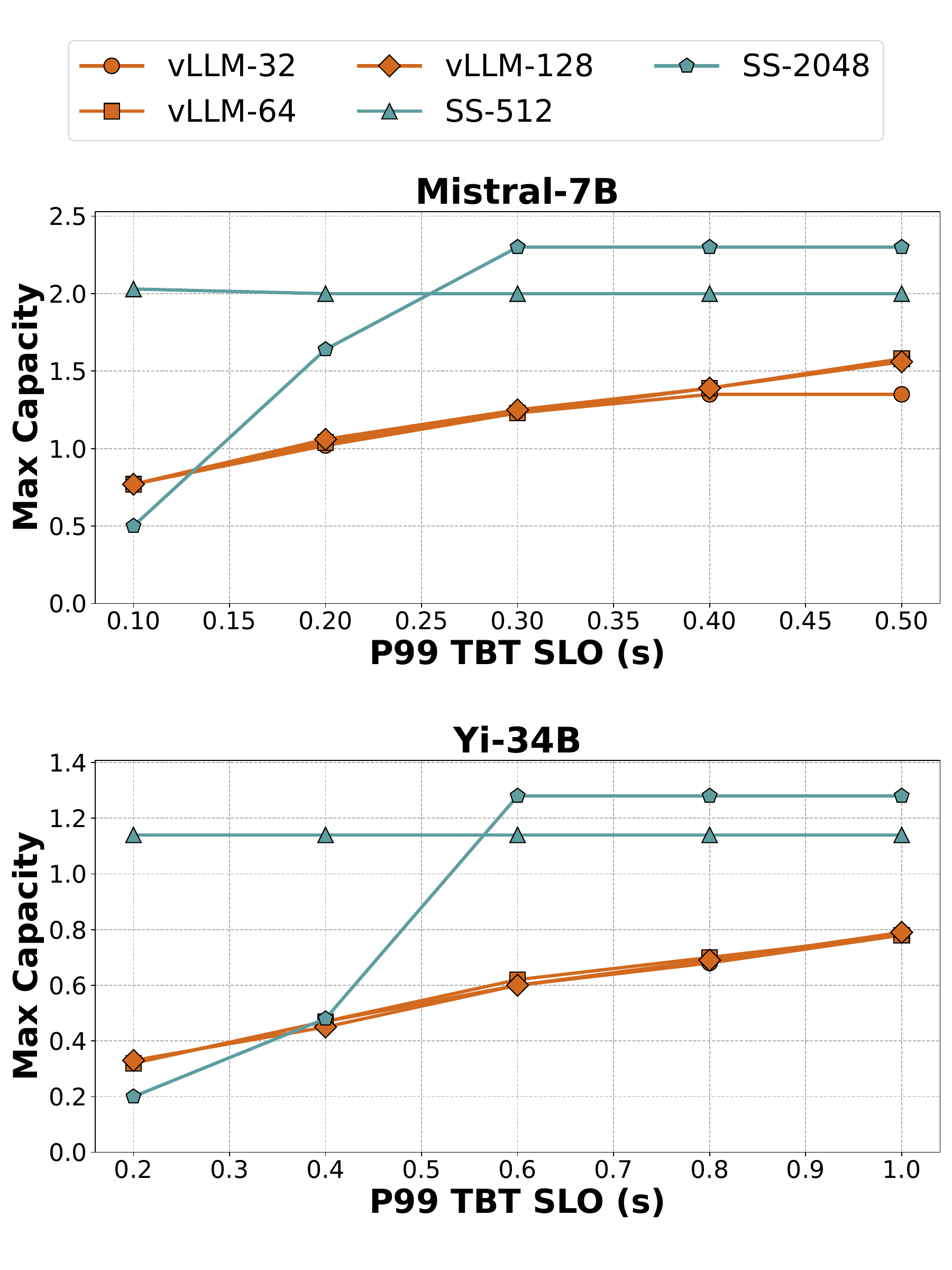}
    \caption{Latency -- Throughput tradeoff in vLLM and \sysname for \mistral and \yi models on \sharegpt dataset. We evaluate vLLM with three different max batch sizes of 32, 64 and 128. For \sysname, we consider token budget of 512 and 2048 with max batch size of 128.
    \sysname delivers 3.5\myx higher capacity under stringent SLOs for \yi using \Hbatch.}
    \label{fig:eval:lateny-tput-tradeoff}
\end{figure}

To fully understand the throughput-latency tradeoff in LLM serving systems, we vary the P99 TBT latency SLO and observe the impact on system capacity for vLLM and \sysname. ~\autoref{fig:eval:lateny-tput-tradeoff} shows the results for \mistral and \yi models with five different SLO values for the \sharegpt dataset.

We evaluate vLLM with three different batch sizes in an attempt to navigate the latency-throughput trade-off as prescribed by Yu et al. \cite{orca}. The maximum capacity of vLLM gets capped due to generation stalls under stringent TBT SLOs. Notably, the capacity of vLLM remains largely identical for all the three batch size settings. This implies that even though PagedAttention enables large batch sizes with efficient memory management -- in practical situations with latency constraints, vLLM cannot leverage the large batch size due to the steep latency-throughput tradeoff made by it's \prefillprio scheduler.

On the other hand, the latency-throughput tradeoff in \sysname can be precisely controlled by varying the token budget. \sysname achieves 3.5\myx higher capacity compared to vLLM under strict SLO (100ms, \mistral) using a small token budget of 512. For scenarios with more relaxed SLO constraints, picking a larger token budget of 2048 allows \sysname to operate more efficiently resulting in 1.65\myx higher capacity compared to vLLM (1s, \yi).

\subsection{Making Pipeline Parallel Viable}
\label{sec-eval-pp}

We now show that \sysname makes it feasible to efficiently serve LLM inference across commodity networks with efficient pipeline parallelism. For these experiments, we run Falcon-180B over two nodes, each with four A100 GPUs, connected over 100 Gbps Ethernet. We evaluate model capacity under three configurations: vLLM with 8-way TP, vLLM with our pipeline-parallel implementation and \sysname with pipeline-parallel. For PP configurations, we do 4-way TP within node and 2-way PP across nodes.

\begin{figure}[t!]
\centering
\begin{subfigure}{0.235\textwidth}
    \includegraphics[width=\textwidth]{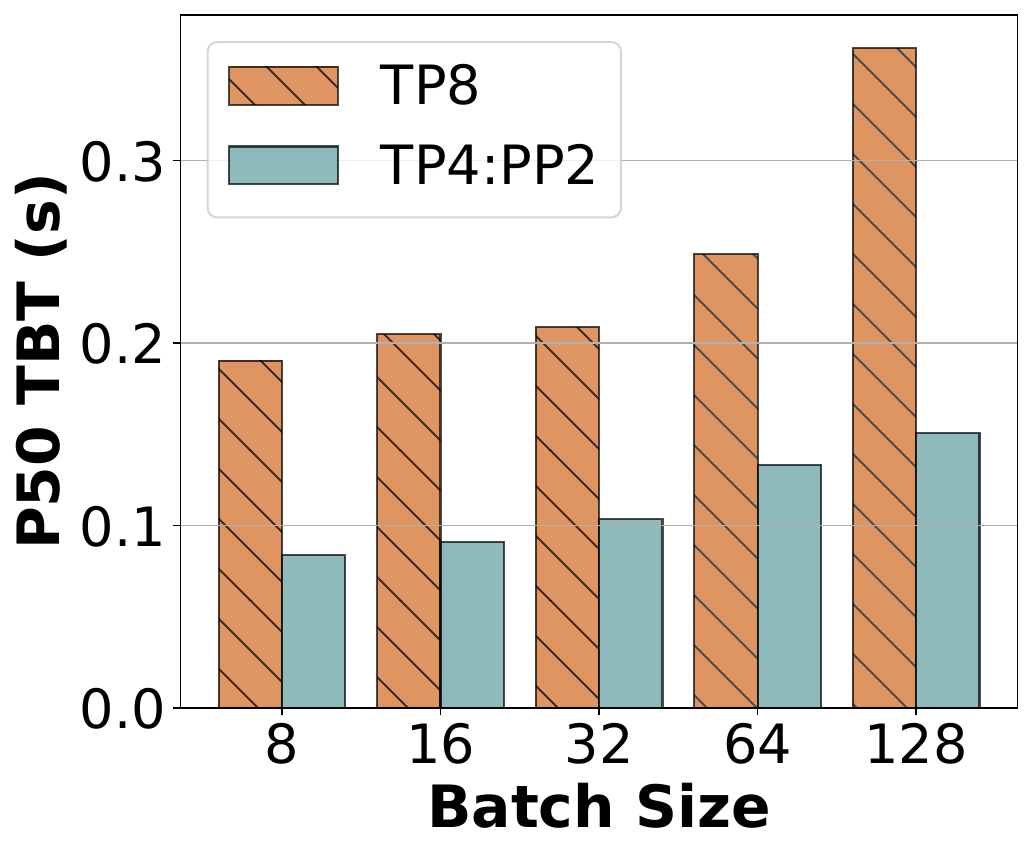}
    \caption{TBT (Falcon-180B).}
    \label{fig:eval:tp-latency}
\end{subfigure}
\begin{subfigure}{0.235\textwidth}
    \includegraphics[width=\textwidth]{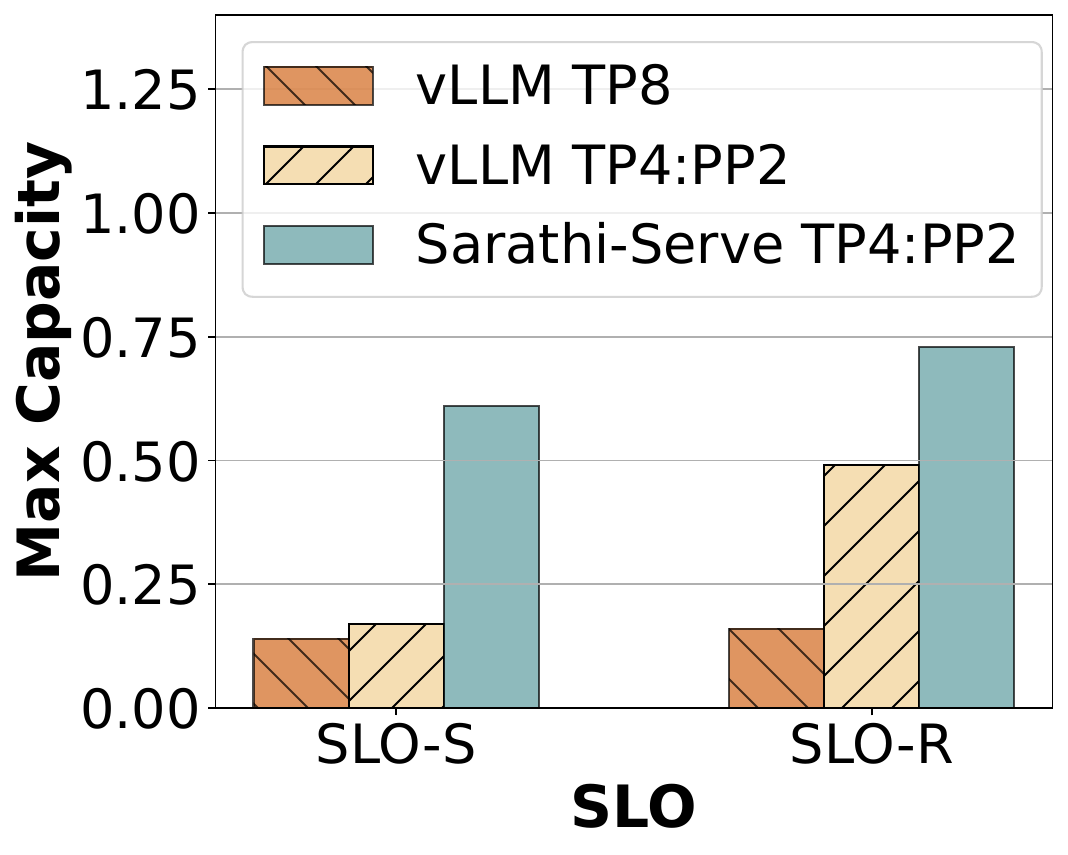}
    \caption{Capacity (Falcon-180B).}
    \label{fig:eval:pp-tp-capacity}
\end{subfigure}
    \caption{TP scales poorly across nodes. (a) Median TBT for decode-only batches: cross node TP increases median TBT by more than $2\times$ compared to a 4-way TP within node and PP across nodes. (b) Capacity under strict (SLO-S) and
relaxed (SLO-R) latency SLOs: \sysname increases Falcon-180B's serving capacity by $4.3\times$ and $3.6\times$ over vLLM's TP-only and hybrid-parallel configurations under strict SLOs.}
    \label{fig:enter-label}
\end{figure}

\autoref{fig:eval:tp-latency} shows the latency for decode-only batches for Falcon-180B with purely tensor parallel TP-8 deployment compared to a TP-4 PP-2 hybrid parallel configuration. We observe that the median latency for tensor parallelism is $\sim 2$\myx higher than pipeline parallelism. This is because TP incurs high communication overhead due to cross-node all-reduces.

\autoref{fig:eval:pp-tp-capacity} shows the capacity for tensor and hybrid parallel configurations for Falcon-180B on \sharegpt dataset. Note that unlike the hybrid parallel configuration, TP achieves low capacity even under the \relaxed \slo due to high latency. Even though vLLM can support a fairly high load with hybrid parallelism under \relaxed \slo, it's performance drops sharply under the \strict regime due to pipeline bubbles. \sysname on the other hand, leverages \chunking to reduce the variation in the execution time between microbatches to avoid pipeline bubbles, resulting in a 1.48\myx increase in capacity under \relaxed \slos and 3.6\myx increase in capacity under \strict \slos.

\subsection{Ablation Study}
\label{sec-eval-ablation}

In this subsection, we conduct an ablation study on different aspects on \sysname. In particular, we are interested in answering the following two questions: 1) what is the effect of chunking on prefill throughput, and 2) analyzing the impact of hybrid-batching and chunking on latency. While we provide results only for a few experiments in this section, all the trends discussed below are consistent across various model-hardware combinations.

\subsubsection{Overhead of \chunking}
\label{sec-eval-ablation1}

\begin{figure}[t!]
    \centering        \includegraphics[width=0.38\textwidth]{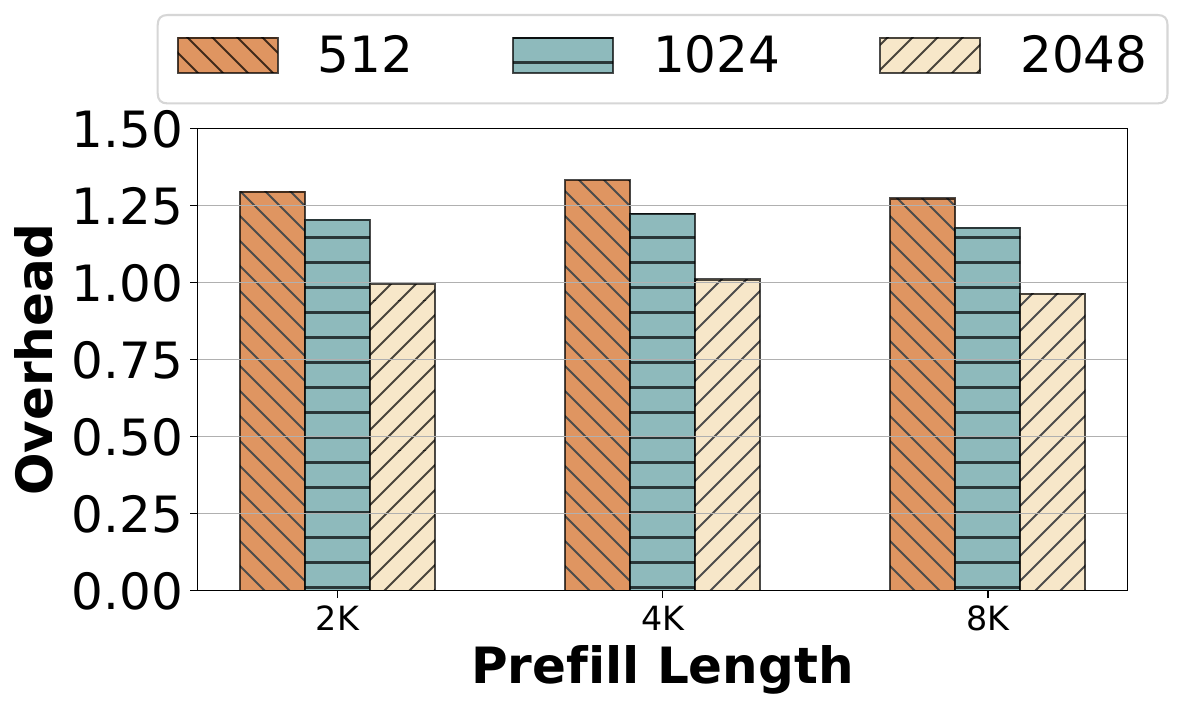}
    \caption{Overhead of \chunking in prefill computation for Yi-34B (TP-2) normalized to the cost of no-chunking, shown for various prompt lengths using chunk lengths of 512, 1024 and 2048.}
    \label{fig:eval:ablation:overheads}
\end{figure}

\autoref{fig:eval:ablation:overheads} shows how much overhead chunking adds in \yi{} -- on overall prefill runtime. As expected, smaller chunks introduce higher overhead as shown by the gradually decreasing bar heights in~\autoref{fig:eval:ablation:overheads}. However, even with the smallest chunk size of 512, we observe a moderate overhead of at most $\sim$25\%. Whereas with the larger token budget of 2048, chunked prefills have almost negligible overhead. %

\begin{table}[t]
\centering
\setlength{\tabcolsep}{2pt}
\scalebox{0.85}{
\begin{tabular}{c|cc|cc}
\textbf{Scheduler} & \multicolumn{2}{c|}{\textbf{\sharegpt}} & \multicolumn{2}{c}{\textbf{\arxivsum}} \\
\cline{2-3} \cline{4-5}
& P50 TTFT & P99 TBT & P50 TTFT & P99 TBT \\ \toprule
\dcoel{}\textit{-only} & 0.53 & 0.68 & 3.78 & 1.38 \\
\chunking{}\textit{-only} & 1.04 & 0.17 & 5.38 & 0.20 \\
\sysname (combined) & 0.76 & 0.14 & 3.90 & 0.17 \\
\bottomrule
\end{tabular}}
\caption{TTFT and TBT latency measured in seconds for \dcoel and \chunking used in isolation as well as when they are used in tandem, evaluated over 128 requests for Yi-34B running on two A100s with a token budget of 1024. By using both \dcoel and \chunking, \sysname is able to lower both TTFT and TBT.}
\label{table:eval:attribution}
\end{table}

\subsubsection{Impact of individual techniques}
\label{sec-eval-ablation2}
Finally, \autoref{table:eval:attribution} shows the TTFT and TBT latency with each component of  \sysname evaluated in isolation \ie \chunking-only, \dcoel-only (mixed batches with both prefill and decode requests) and when they are used in tandem. These results show that the two techniques work best together: \chunking-only increases TTFT as prefill chunks are slightly inefficient whereas \dcoel-only increases TBT because long prefills can still create generation stalls. When used together, \sysname improves performance along both dimensions.

\vspace{1em}
\section{Related Work}
\label{section:relatedwork}

\noindent{\bf Model serving systems:} Systems such as Clipper~\cite{clipper}, TensorFlow-Serving~\cite{tensorflowserving}, Clockwork~\cite{clockwork} and BatchMaker~\cite{batchmakereurosys} study various placement, caching and batching strategies for model serving. However, these systems fail to address the challenges of auto-regressive transformer inference. More recently, systems such as Orca~\cite{orca}, vLLM~\cite{vllmsosp}, FlexGen~\cite{flexgen}, FasterTransformers~\cite{fastertransformer}, LightSeq~\cite{lightseq}, and TurboTransformers~\cite{turbotransformers}  propose domain-specific optimizations for transformer inference. FlexGen~\cite{flexgen} optimizes LLM inference for throughput in resource-constrained offline scenarios i.e., it is not suitable for online serving. FastServe~\cite{fastserve} proposed a preemptive scheduling framework for LLM inference to minimize the job completion times. We present a detailed comparison with Orca and vLLM as they represent the state-of-the-art in LLM inference.

Another approach that has emerged recently is to disaggregate the prefill and decode phases on different replicas as proposed in SplitWise, DistServe and TetriInfer ~\cite{patel2023splitwise,distserve2024,tetriinfer}. These solutions can entirely eliminate the interference between prefills and decodes. However, disaggregation requires migrating the KV cache of \textit{each} request upon the completion of its prefill phase which could be challenging in the absence of high-bandwidth interconnects between different replicas. In addition, this approach also under-utilizes the GPU memory capacity of the prefill replicas i.e., only the decode replicas are responsible for storing the KV cache. On the positive side, disaggregated approaches can execute prefills with maximum efficiency (and therefore yield better TTFT) unlike chunked prefills that are somewhat slower than full prefills. We leave a quantitative comparison between \sysname and disaggregation-based solutions for future work.

Recently, Sheng et al. \cite{sheng2023fairness} proposed modification to \ils algorithm to ensure fairness among clients in a multi-tenant environment. FastServe~\cite{fastserve} uses a preemption based scheduling mechanism to mitigate head-of-the-line blocking. Such algorithmic optimizations are complimentary to our approach and can benefit from lower prefill-decode interference enabled by \sysname. Another recent system, APIServe \cite{abhyankar2024apiserve} adopted chunked prefills from \sarathi to utilize wasted compute in decode batches for ahead-of-time prefill recomputation for multi-turn API serving.

\noindent{\bf Improving GPU utilization for transformers:} Recent works have proposed various optimizations to improve the hardware utilization for transformers. FasterTransformer uses model-specific GPU kernel implementations. CocoNet~\cite{coconetasplos} and ~\cite{googleoverlap} aim to overlap compute with communication to improve GPU utilization: these techniques are specially useful while using a high degree of tensor-parallel for distributed models where communication time can dominate compute. Further, the cost of computing self-attention grows quadratically with sequence length and hence can become significant for long contexts. \cite{self:attn:on2:memory, flashattention, flashattention2} have proposed various techniques to minimize the memory bottlenecks of self-attention with careful tiling and work partitioning. In addition, various parallelization strategies have been explore to optimize model placement. These techniques are orthogonal to \sysname.

\noindent{\bf Model optimizations:} A significant body of work around model innovations has attempted to address the shortcomings of  transformer-based language models or to take the next leap forward in model architectures, beyond transformers. For example, multi-query attention~\cite{multiqueryattention} shares the same keys and values across all the attention heads to reduce the size of the KV-cache, allowing to fit a larger batch size on the GPUs. Several recent works have also shown that the model sizes can be compressed significantly using quantization~\cite{smoothquant,gptq,qlora,llmint8}. Mixture-of-expert models are aimed primarily at reducing the number of model parameters that get activated in an iteration~\cite{largescalemoe,moeatc23,moedeployment}. More recently, retentive networks have been proposed as a successor to transformers~\cite{retnet}. In contrast, we focus on addressing the performance issues of popular transformer models from a GPU's perspective. 

\section{Conclusion}
\label{sec-conclusion}

Optimizing LLM inference for high throughput and low latency is desirable but challenging. We presented a broad characterization of existing LLM inference schedulers by dividing them into two categories -- \prefillprio and \decodeprio. In general, we argue that the former category is better at optimizing throughput whereas the latter is better at optimizing TBT latency. However, none of them is ideal when optimizing throughput and latency are both important. 

To address this tradeoff, we introduce \sysname --- a system that instantiates a novel approach comprised of  \chunking and \hbatch. \sysname chunks input prompts into smaller units of work to create stall-free schedules. This way, \sysname can add new requests in a running batch without pausing ongoing decodes. Our evaluation shows that \sysname improves the serving capacity of \mistral by up to 2.6\myx on a single A100 GPU and up to 5.6\myx for \falcon on 8 A100 GPUs.

\section{Acknowledgement}
We would like to thank OSDI reviewers and our shepherd for their insightful feedback. This research is partly supported by GT Cloud Hub, under the auspices of the Institute
for Data Engineering and Science (IDEaS), with funding from Microsoft, and the Center for Research into Novel Compute Hierarchies (CRNCH) at Georgia Tech.

\phantomsection
\label{EndOfPaper}
\bibliographystyle{plain}
\bibliography{all}

\appendix
\section{Artifact Appendix}

\subsection*{Abstract}
Our open source artifact is available on \href{https://github.com/microsoft/sarathi-serve}{GitHub}. This repository contains our implementation of \sysname as well as the harnesses and scripts for
running and plotting the experiments described in this paper.

This repository originally started as a fork of the vLLM project. \sysname is a  lightweight high-performance research prototype and doesn't have complete feature parity with open-source vLLM. We have only retained the most critical features and adopted the codebase for faster research iterations.

\subsection*{Scope}

This artifact allows the readers to validate the claims made
in the \sysname paper (the figures) and provides a means to replicate the experiments described. The artifact
can be used to set up the necessary environment, execute the
main results, and perform microbenchmarks, thus providing a comprehensive understanding of the key claims in \sysname. %

\subsection*{Contents}

The repository is structured as follows, the primary source code for the system is contained in directory \textit{/sarathi}. The implementations for custom CUDA kernels are within the \textit{/csrc} directory. All the scripts to reproduce the experiments are in \textit{/osdi-experiments} and finally, the trace files used for the experiments are stored in \textit{/data}.

\subsection*{Hosting}
You can obtain our artifacts from GitHub:
\href{https://github.com/microsoft/sarathi-serve}{GitHub}. The main branch of the Github repository is actively updated, but we will maintain clear and accessible instructions about our artifacts in an easily identifiable
README file. All the detailed instructions and README files to reproduce the experiments in the OSDI paper are available in the branch \textit{osdi-sarathi-serve}.

\subsection*{Requirements}

\sysname has been tested with CUDA 12.1 on A100 and A40 GPUs. The specific GPU SKUs on which the experiments were performed and the parallelism strategies used are clearly explained in the README corresponding to the figures in the artifact, for ease of reproducibility.

\end{document}